\documentclass[11pt,a4paper]{article}
\pdfoutput=1
\usepackage[hmargin=3.0cm,vmargin=3.0cm]{geometry}

\usepackage[table]{xcolor}
\usepackage{booktabs}  % tables: \toprule \midrule \bottomrule instead of \hline
\usepackage{graphicx}
\usepackage{caption,subcaption}
\usepackage{mathtools}
\usepackage{amsfonts}
\usepackage[inline]{enumitem}  % inline enumeration
\usepackage{hyperref}
\usepackage[noabbrev]{cleveref}  % include the word "Figure" automatically
\usepackage{authblk}            % author affiliation etc
\hypersetup{
colorlinks=true,
citecolor=red
}

\usepackage[group-separator={,}]{siunitx}  % 1,234,567
\usepackage{multicol}
\usepackage[numbers,sort&compress]{natbib}  % plainnat bibliography style

\usepackage[space]{grffile}  % includegraphics filename with spaces
\usepackage[pagewise]{lineno}
\usepackage{algorithm}
\usepackage{algpseudocode}

\usepackage[bottom]{footmisc}  % put footnotes under possible floats

\newcommand\PP{\mathbb{P}}
\newcommand\QQ{\mathbb{Q}}
\newcommand\RR{\mathbb{R}}
\newcommand\EE{\mathop{\mathbb{E}}}
\newcommand\zz{\mathbf{z}}

\newcommand\yy{\mathbf{y}}
\newcommand\ww{\mathbf{w}}

\newcommand\manL{\mathcal{L}}
\newcommand\manN{\mathcal{N}}

\newcommand\manU{\mathcal{U}}
\newcommand\manD{\mathcal{D}}
\newcommand\manO{\mathcal{O}}

\DeclareMathOperator*{\argmax}{arg\,max}
\DeclareMathOperator*{\argmin}{arg\,min}

\begin{document}%\sloppy

\title{Parametric generation of conditional geological realizations using generative neural networks\footnote{Published in Computational Geosciences (2019). DOI: \href{https://doi.org/10.1007/s10596-019-09850-7}{10.1007/s10596-019-09850-7}}
}

\author[]{Shing Chan\footnote{Corresponding author.\\ E-mail addresses: \texttt{sc41@hw.ac.uk} (Shing Chan), \texttt{a.elsheikh@hw.ac.uk} (Ahmed H. Elsheikh).} }
\author[]{Ahmed H. Elsheikh}
\affil[]{Heriot-Watt University, Edinburgh, UK}

\maketitle

\begin{abstract}
Deep learning techniques are increasingly being considered for geological applications where -- much like in computer vision -- the challenges are characterized by high-dimensional spatial data dominated by multipoint statistics.
In particular, a novel technique called \emph{generative adversarial networks} has been recently studied for geological parametrization and synthesis, obtaining very impressive results that are at least qualitatively competitive with previous methods.
The method obtains a neural network parametrization of the geology -- so-called a \emph{generator} --  that is capable of reproducing very complex geological patterns with dimensionality reduction of several orders of magnitude.
Subsequent works have addressed the conditioning task, i.e. using the generator to generate realizations honoring spatial observations (hard data). The current approaches, however, do not provide a parametrization of the conditional generation process.
In this work, we propose a method to obtain a parametrization for direct generation of conditional realizations.
The main idea is to simply extend the existing generator network by stacking a second \emph{inference network} that learns to perform the conditioning.
This inference network is a neural network trained to sample a posterior distribution derived using a Bayesian formulation of the conditioning task.
The resulting extended neural network thus provides the conditional parametrization.
Our method is assessed on a benchmark image of binary channelized subsurface, obtaining very promising results for a wide variety of conditioning configurations.
\end{abstract}

\section{Introduction}\label{sec:intro}
% \lhead{\emph{Introduction}}

The large scale nature of geological models makes reservoir simulation an expensive task,
prompting numerous works on parametrization methods that can preserve complex geological characteristics required for accurate flow modeling.
A wide variety of methods exist including
zonation~\citep{jacquard1965,jahns1966}, PCA-based
methods~\citep{sarma2008kernel,ma2011kernel,vo2016regularized}, SVD
methods~\citep{shirangi2016improved,tavakoli2011monte},
discrete cosine transform~\citep{jafarpour2009,jafarpour2010},
level set methods~\citep{moreno2007stochastic,dorn2008,chang20108011}, and
dictionary learning~\citep{khaninezhad2012sparse1,khaninezhad2012sparse2}.
Very recently, a new method from the machine learning community called
\emph{generative adversarial networks}~\citep{goodfellow2014generative} has been
investigated~\citep{mosser2017reconstruction,mosser2017stochastic,chan2017parametrization,laloy2018training,dupont2018generating,mosser2018conditioning,chan2019parametrization}
for the purpose of parametrization, reconstruction, and synthesis of geological
properties; obtaining very competitive results in the visual quality of the generated images compared to previous methods.
This adds to the recent trend in applying machine learning techniques~\citep{marccais2017prospective,kani2017dr,klie2015physics,stanev2018identification,sun2017new,zhu2018bayesian,valera2017machine}
to leverage rapid advances in the field as well as the increasing availability of data and computational resources that enable these techniques to be effective.

Generative adversarial networks (GAN) is a novel technique for training a neural network to sample from a distribution that is unknown and intractable, by only using a dataset of realizations from this distribution. The result is a neural network parametrization called a \emph{generator}, which is capable of generating new realizations from the target distribution -- in our case, geological images -- using a very efficient representation. Recent works show that using the generator to parametrize the geology is very effective in preserving high-order flow statistics~\citep{chan2017parametrization,chan2019parametrization}, two-point spatial statistics~\citep{mosser2017reconstruction,laloy2018training} and morphology~\citep{mosser2017reconstruction}, all while achieving dimensionality reduction of several orders of magnitude.

Subsequent works on GAN focused on the problem of conditioning the generator: given a generator trained on unconditional realizations, the task is to generate realizations conditioned to spatial observations (hard data).
In~\citep{dupont2018generating,mosser2018conditioning}, an image inpainting technique was used which adopts a sampling by optimization approach, i.e. it requires solving an optimization problem for each conditional realization that is generated.
The method obtained very good results -- in particular,~\citep{dupont2018generating} reported superior performance in many aspects compared to standard geomodeling tools.
However, sampling by optimization can be expensive if realizations need to be continuously generated during deployment, e.g. for history matching or uncertainty quantification.
An alternative approach was presented in~\citep{laloy2018training}, where the authors addressed conditioning using a Bayesian framework and performed Markov chain Monte Carlo to sample conditional realizations.
Neither of these approaches, however, provides a parametrization for the conditional sampling process.
As the authors in~\citep{laloy2018training,dupont2018generating} express, it is of interest to obtain such parametrization to directly sample conditional realizations without running optimizations or Monte Carlo methods.

In this work, we propose a method to obtain a parametrization to directly sample conditional realizations.
The main idea is to simply extend the existing generator network by stacking a second \emph{inference network} that performs the conditioning.
This inference network is a neural network trained to sample a posterior distribution, derived using a Bayesian formulation of the conditioning task.
The resulting extended neural network thus provides the conditional parametrization, hence direct conditional sampling can be done very efficiently. We assess our method on the benchmark image of~\citet{strebelle2001reservoir}, finding positive results for a wide variety of conditioning configurations.

Note that although previous works~\citep{mosser2017reconstruction,laloy2018training,dupont2018generating} study applications of GAN mainly in the context of geomodeling and multipoint geostatistical simulations, here we emphasize on the effectiveness of GAN -- and neural networks in general -- for parametrization and dimensionality reduction, highlighting their ability to learn efficient representations for complex and high-dimensional data.
The rest of this work is organized as follows:
In~\Cref{sec:background}, we describe parametrization using generative adversarial networks, and the Bayesian formulation of the conditioning problem.
We introduce our method in~\Cref{sec:methodology} where we describe how the inference network is obtained.
In~\Cref{sec:results}, we show results for unconditional and conditional parametrization of binary channelized subsurface images.
We discuss related work in~\Cref{sec:related} including other alternatives to train the inference network, and conclude our work in~\Cref{sec:conclusion}.

\section{Background} \label{sec:background}
In this section, we discuss the importance of parametrization for subsurface simulations (\Cref{sec:parametrization}), we describe generative adversarial networks (\Cref{sec:gan}), and we describe the Bayesian formulation of the conditioning problem (\Cref{sec:conditioning}).

\subsection{Parametrization}\label{sec:parametrization}
\begin{figure}\centering
    \includegraphics[width=.8\textwidth]{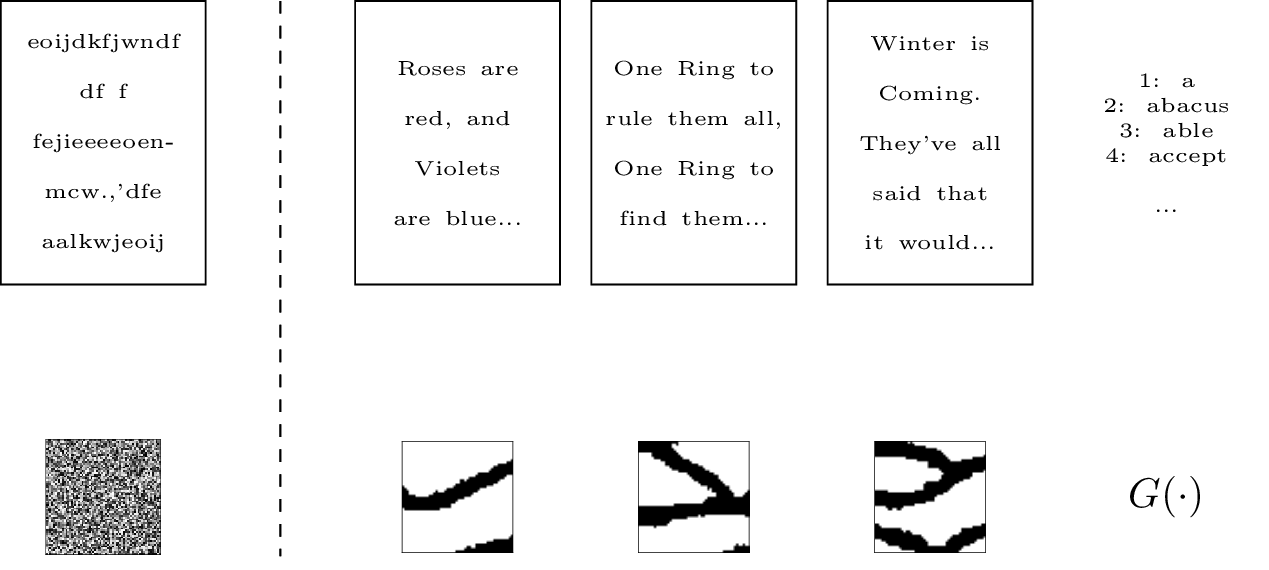}
    \caption{An index of words provides all \emph{plausible} arrangement of letters (top row). Similarly, a geological parametrization provides all \emph{plausible} realizations of the subsurface (bottom row).\label{fig:parametrization}}
\end{figure}

Parametrization is useful in subsurface simulations where the large number of uncertain variables are highly correlated and redundantly represented as a consequence of the grid discretization.
One useful analogy to parametrization is an index of words or a dictionary: Consider the task of inferring the content of a book using only indirect information such as the frequency of letters. A priori, this task would need to consider any possible arrangement of letters however implausible (top row, left of~\Cref{fig:parametrization}). On the other hand, since most books consist of words, we know that most arrangements are unlikely and can be quickly discarded. The task, although still difficult,  is suddenly much easier with the inclusion of this prior information via an index of words (top row, right of~\Cref{fig:parametrization}).
Likewise, consider the task of inferring the subsurface from indirect information such as the oil production history.
Without any other information, attempting to deliberately model the subsurface to match the production history would almost certainly result in unrealistic images (bottom row, left of~\Cref{fig:parametrization}).
On the other hand, we know that real subsurface images are not completely random but instead tend to exhibit clear spatial correlations. By using a suitable parametrization of the subsurface, we can embed this information and narrow our search to only the plausible realizations (bottom row, right of~\Cref{fig:parametrization}), thus reducing the number of simulations required in uncertainty quantification and inversion problems.

Let the random vector $\yy\in\RR^{n_y}$ represent plausible subsurface images. Parametrization aims to construct a well-behaved function $G\colon\RR^{n_z}\to\RR^{n_y}$ such that $\yy=G(\zz)$ where $\zz\in\RR^{n_z}$ (normally $n_z\ll n_y$) is a \emph{latent} random vector with known pre-defined distribution (for example, $\zz\sim\mathcal{N}(\mathbf{0},\mathbf{I})$).
Generally, strictly achieving $\yy=G(\zz)$ for complex and high-dimensional $\yy$ is hard, hence many methods settle for replicating simple statistics of $\yy$ such as the mean and covariance.
For example, in a parametrization based on principal component analysis, $G$ is an affine transformation

$$G(\zz) = A\zz + b$$
where $A,\;b$ are fitted so that $G(\zz),\;\zz\sim\mathcal{N}(\mathbf{0},\mathbf{I})$ preserves the sample mean and covariance estimated from an available dataset $\{y_1,\cdots,y_n\}$ of realizations of $\yy$.
Note that for nature this parametrization is often too simplistic, resulting in unrealistic realizations that are overly smooth in practice.

In this work, we use a parametrization based on deep neural networks:

\begin{equation}\label{eq:dnn}
    G(\zz) = f_l \circ f_{l-1} \circ \cdots \circ f_1(\zz),\quad f_i(x) = \sigma_i(A_ix+b_i)
\end{equation}
where $\circ$ denotes composition ($f_2\circ f_1(x) = f_2(f_1(x))$), and $\sigma_i$ denotes a component-wise non-linearity\footnote{The non-linearity adds expressivity, otherwise the composition reduces to an affine transformation. Typical choices include $\tanh(x)$, $\max(0,x)$ and $\operatorname{sigmoid}(x)$.}. This is motivated by the high expressive power of deep neural networks as it is now evident from the state-of-the-art results in computer vision (see e.g.~\citep{brock2018large, karras2017progressive} for recent examples). In addition to the more flexible parametrization, instead of training the weights $A_i, b_i$ to preserve only mean and covariance as in principal component analysis, we leverage once more the expressive power of neural networks and let a second neural network learn and decide the relevant statistics directly from the dataset. This is possible due to a recent technique called generative adversarial networks, described below in~\Cref{sec:gan}.

\subsection{Generative adversarial networks}\label{sec:gan}

We use generative adversarial networks to obtain the (unconditional) parametrization of the geology.
This method can be used to obtain a parametrization of a general random vector given a dataset of its realizations.
Let the random vector $\yy\in\RR^{n_y}$ represent the uncertain subsurface property of interest,
where $n_y$ is very large (e.g. permeability discretized by the simulation grid).
This random vector follows a distribution $\yy\sim \PP_y$ that is unknown and
possibly intractable (e.g. distribution of plausible channelized permeability images).
Instead, we are only given a dataset of realizations $\{y_1,\cdots,y_N\}$ of the
random vector (e.g. a set of permeability realizations deemed representative
of the area under study).
Using this training dataset, we aim to find a parametrization for $\yy$:
Consider now a \emph{latent} random vector $\zz\in\RR^{n_z}$ with $n_z \ll n_y$
and $\zz\sim p_z$ where $p_z$ is manually chosen to be easy to sample from
(e.g. a multivariate normal or uniform distribution); and a neural network
$G_\theta\colon \RR^{n_z}\to\RR^{n_y}$, that we call a \emph{generator},
where $\theta$ denotes the weights of the neural network. We aim to determine $\theta$ so that $\yy = G_\theta(\zz)$.
In other words, let $\PP_\theta$ denote the distribution induced by $G_\theta$ (i.e. $G_\theta(\zz)\sim\PP_\theta$), which depends on $\theta$;
the goal is to determine $\theta$ so that  $\PP_\theta = \PP_y$.

A difficulty with the problem statement above is that $\PP_y$ is completely unknown (we only have realizations of $\yy$) and $\PP_\theta$ is unknown and intractable (even if $p_z$ is simple, $G_\theta$ is a neural network with several non-linearities).
On the other hand, sampling from these distributions is easy: For $\PP_y$, we ``sample'' by drawing realizations from the training set, assuming the set is large enough to be representative. For $\PP_\theta$, we simply sample $\zz\sim p_z$ and evaluate $G_\theta(\zz)$.
We therefore have two distributions that we can sample from but we cannot model analytically, and yet we need to optimize $\PP_\theta$ to approximate $\PP_y$. Informally, we need to teach the generator $G_\theta$ to generate \emph{plausible} realizations.

\begin{figure}\centering
    \includegraphics[width=.6\textwidth]{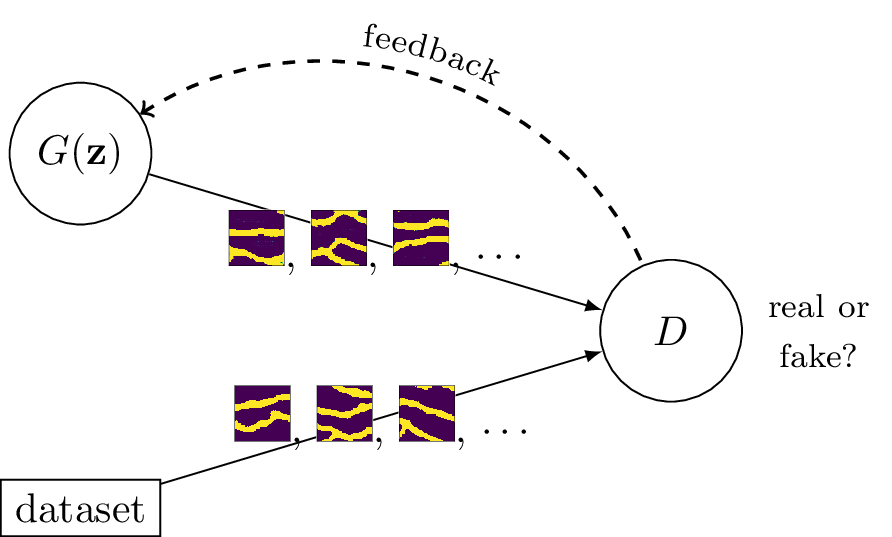}
    \caption{Generative adversarial networks.\label{fig:gan}}
\end{figure}

Following this observation, the seminal work in~\citep{goodfellow2014generative} (see also~\citep{schmidhuber1992learning}) introduces the idea of using a classifier neural network $D_\psi\colon \RR^{n_y}\to [0,1]$ called a \emph{discriminator}, with weights $\psi$, to assess the \emph{plausibility} of generated realizations.
The discriminator $D_\psi$ is trained to distinguish between ``fake'' (from generator) and ``real'' (from training dataset) realizations, and it essentially outputs a probability estimate. The aim of the generator is then to fool the discriminator (see~\Cref{fig:gan}),
hence the discriminator and the generator are adversaries.
The discriminator is trained to solve a binary classification problem by maximizing the following loss:

\begin{align}
  \manL(\psi,\theta)  &\coloneqq \EE_{\yy\sim\PP_y}{\log D_{\psi}(\yy)} + \EE_{\tilde{\yy}\sim\PP_\theta}{\log(1-D_{\psi}(\tilde{\yy}))} \label{eq:ganloss}  \\
  &\approx \frac{1}{M}\sum^M_{i=1}{\log D_{\psi}(y_i)} + \frac{1}{M}\sum^M_{i=1}{\log(1-D_{\psi}(G_\theta(z_i))} \nonumber
\end{align}
which is in essence a binary classification score.
The expectations in the expression above are approximated by taking a batch of $M\leq N$ realizations from the
training set for the first term, and sampling $M$ realizations $z_1,\cdots,z_M$ from $p_z$ for the second term.

The generator on the other hand is trained to minimize the \emph{same} loss. Thus, an
adversarial game is created where $G$ and $D$ optimize the loss in opposite
directions,
\begin{equation}
  \label{eq:minmax}
  \min_\theta \max_\psi {\manL(\psi,\theta)}
\end{equation}
In practice, this optimization is performed alternately using
stochastic gradient descent, where the gradients with respect to $\theta$ and $\psi$
are obtained using automatic differentiation algorithms. The equilibrium is
reached when $G$ effectively learns to approximate $\PP_y$ and $D$ is
$\frac{1}{2}$ in the support of $\PP_y$ (coin toss scenario). It is shown in~\citep{goodfellow2014generative} that in the infinite capacity setting, the process minimizes
the Jensen-Shannon divergence between $\PP_\theta$ and $\PP_y$.
Once trained, we can discard the discriminator and keep the generator as our parametrization.

Note that the method is very general and directly applicable in practice to all types of geological models including multi-facies and multimodal geology, since minimal assumptions are imposed on $\PP_y$ and $\PP_\theta$ as we do not need to model them explicitly. We only require realizations $\{y_1,\cdots,y_N\}$ from the unknown target distribution $\PP_y$ and the discriminator is in charge of inferring it from the realizations.

\paragraph*{Variations of GAN}
Stability issues with the original formulation of GAN led to numerous works
to improve and generalize the method (see e.g.
\citep{radford2015unsupervised,salimans2016improved,arjovsky2017towards,arora2017generalization}
and references therein). One line of research generalizes GAN in the framework
of integral probability metrics~\citep{muller1997integral}: Given two distributions
$\PP$ and $\QQ$, and a class of real valued functions $\manD$, an integral
probability metric measures the discrepancy between $\PP$ and $\QQ$ as follows:

\begin{equation}\label{eq:ipm}
  \operatorname{d}_{\manD}(\PP,\QQ) = \sup_{D\in\manD}\{\EE_{\yy\sim\PP} D(\yy) - \EE_{\tilde{\yy}\sim\QQ} D(\tilde{\yy})\} \nonumber
\end{equation}
Note the slight similarity with~\Cref{eq:ganloss}. In comparison, this new formulation\footnote{Actually, this formulation precedes GAN by almost two decades~\citep{muller1997integral}, although it is introduced in a different context within probability theory. The connection was drawn recently and led to the numerous works mentioned.}
drops the logarithms and performs the optimization within a class $\manD\ni D$ that may be more general, i.e. not necessarily limited to classifier functions.
The choice of $\manD$ is
important and leads to different flavors of GAN.
For example, when $\manD$ is a ball in a Reproducing Kernel Hilbert Space, $\operatorname{d}_\manD$ is the Maximum Mean Discrepancy (MMD GAN)~\citep{gretton2007kernel,dziugaite2015training}.
When $\manD$ is a set of 1-Lipschitz functions, $\operatorname{d}_\manD$ is the Wasserstein distance (WGAN)~\citep{arjovsky2017wasserstein,gulrajani2017improved}.
When $\manD$ is a Lebesgue ball, we obtain Fisher GAN~\citep{mroueh2017fisher}, and when $\manD$ is a Sobolev ball, we obtain Sobolev GAN~\citep{mroueh2017sobolev}.
See~\citep{mroueh2017mcgan,mroueh2017sobolev} for further discussion.
Our unconditional generator is trained using the Wasserstein formulation (see also our recent work~\citep{chan2017parametrization,chan2019parametrization}).

\subsection{Conditioning to observations} \label{sec:conditioning}
Given a pre-trained generator $G$, we aim to generate realizations conditioned to spatial observations (hard data), i.e. find $z$ such that $G(z)$ honors the observations.
Let $d_{\mathrm{obs}}$ denote the
observations and $d(z) = G(z)_{\mathrm{obs}}$ the values at the
observed locations given $G(z)$.
Under the probabilistic framework, we can formulate the problem as finding $z^*$ that maximizes its
posterior probability given observations,
\begin{equation}\label{eq:argmax_posterior}
z^* = \argmax_z{p(z|d_{\mathrm{obs}}})
\end{equation}
From Bayes' rule and applying logarithms,

\begin{align}
  p(z | d_{\mathrm{obs}}) &\propto  p(d_{\mathrm{obs}} | z)p(z) \nonumber \\
  - \log p(z | d_{\mathrm{obs}}) &= - \log p(d_{\mathrm{obs}} | z) - \log p(z) + \mathrm{const.} \nonumber
\end{align}
For the prior $p(z)$, a natural choice is $p_z$ for which the generator has been
trained. In most applications (and in ours), this is the multivariate
standard normal distribution, then $p(z)\propto \exp(-\frac{1}{2}\|z\|^2)$. For the likelihood $p(d_{\mathrm{obs}}|z)$, we take the general
assumption of i.i.d. Gaussian measurement noise, $p(d_{\mathrm{obs}}|z) \propto
\exp(-\frac{1}{2\sigma^2}\|d(z)-d_{\mathrm{obs}}\|^2)$ where $\sigma$ is the
measurement standard deviation. Then the optimization
in~\Cref{eq:argmax_posterior} can be written as

\newcommand\myeq{\stackrel{\mathclap{\mbox{\scalebox{0.5}{($\times 2\lambda$)}}}}{=}}
\begin{equation}
  z^* = \argmin_z \manL(z) \label{eq:optim}
\end{equation}
where

\begin{align}
  \manL(z) &\coloneqq -\log p(z | d_{\mathrm{obs}}) \nonumber \\
  &\myeq \|d(z) - d_{\mathrm{obs}}\|^2 + \lambda\|z\|^2 \nonumber \\
  &= \|G(z)_{\mathrm{obs}} - d_{\mathrm{obs}}\|^2 + \lambda\|z\|^2 \label{eq:lossfn}
\end{align}
where we multiplied everything by $\lambda=\sigma^2$ and discarded the
irrelevant constant.
One way to draw different conditional realizations is to
optimize~\Cref{eq:optim} repeatedly using a local optimizer and different
initial guesses for $z$, as performed
in~\citep{mosser2018conditioning,dupont2018generating}. Another approach is
to sample the full posterior using Markov chain Monte Carlo methods as
performed in~\citep{laloy2018training}.

\section{Conditional generator for geological realizations} \label{sec:methodology}

\begin{figure}\centering
	\includegraphics[width=.7\textwidth]{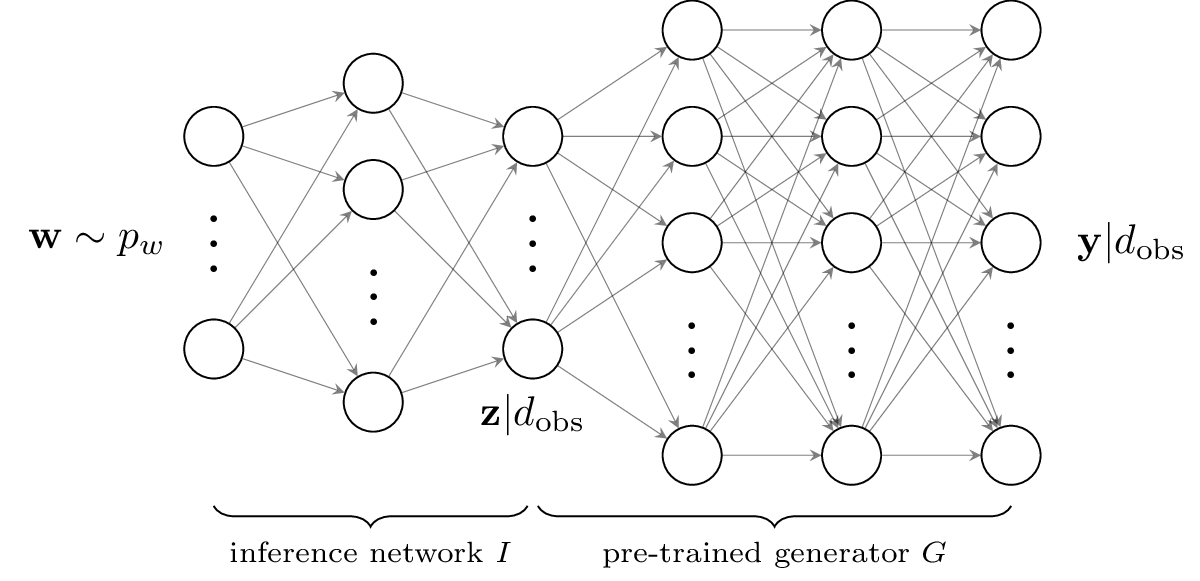}
	\caption{Illustration of methodology, $G\circ I$.\label{fig:methodology}}
\end{figure}

As mentioned in~\Cref{sec:conditioning}, one way to sample multiple realizations conditioned to observations is to solve~\Cref{eq:optim} repeatedly using local optimization with different initial guesses~\citep{dupont2018generating,mosser2018conditioning}.
However, this approach can be expensive if a large number of realizations need to be continuously generated in deployment, e.g. for uncertainty quantification and inversion problems, and it also may not cover the full solution space.
Another approach is to use Markov chain Monte Carlo methods -- assuming the latent vector is of moderate size -- to sample the full posterior distribution~\citep{laloy2018training}.
Neither approach, however, provides a parametrization of the sampling process. That is, we no longer have a functional relationship $\yy_{\mathrm{cond}} = G_{\mathrm{cond}}(\ww),\;\ww\sim p_w$, where $\yy_{\mathrm{cond}}$ denotes the conditional geology and $p_w$ is some fixed distribution.

Here we propose a method to obtain a conditional parametrization for direct and parametric sampling of conditional realizations.
The idea is to extend the existing generator $G\circ I =\colon G_{\mathrm{cond}}$ where $I$ is another neural network -- called the \emph{inference network} -- that performs the conditioning, as illustrated in~\Cref{fig:methodology}.
The inference network is trained to sample the Bayesian posterior $p(z|d_{\mathrm{obs}})$ derived in~\Cref{sec:conditioning}.
Let $I_\phi\colon \RR^{n_w} \to \RR^{n_z}$ where $\phi$ denotes the weights of the neural network to be determined. $I_\phi$ maps from yet another random vector $\ww\in\RR^{n_w},\;\ww\sim p_w$ with manually chosen $p_w$ (we can naturally choose $p_w = p_z$ and $n_z=n_w$), to the conditional latent vector $\zz|d_{\mathrm{obs}}\sim p(z|d_{\mathrm{obs}})$.
Let $q_\phi$ denote the distribution density induced by $I_\phi$, which depends on $\phi$. This density function is unknown and intractable ($I_\phi$ is a neural network with several non-linearities), but is easy to sample from since it only requires sampling $\ww\sim p_w$ and evaluating $I_\phi(\ww)$. The Kullback-Leibler divergence from $p(\cdot|d_{\mathrm{obs}})$ to $q_\phi$ gives us

\begin{align}\label{eq:kl}
  \operatorname{D_{KL}}(q_\phi \parallel p(\cdot|d_{\mathrm{obs}})) &= \EE_{\zz\sim q_\phi} \log \frac{q_\phi(\zz)}{p(\zz|d_{\mathrm{obs}})}\nonumber \\
  &= \EE_{\zz\sim q_\phi}  -\log p(\zz|d_{\mathrm{obs}}) + \EE_{\zz\sim q_\phi} \log q_\phi(\zz) \nonumber \\
  &= \EE_{\zz\sim q_\phi} \manL(\zz) + \EE_{\zz\sim q_\phi} \log q_\phi(\zz) + \mathrm{const.}
\end{align}
The first term is the expected loss under the induced distribution $q_\phi$, with the loss defined in~\Cref{eq:lossfn}. It can be approximated as

\begin{equation}
  \EE_{\zz\sim q_\phi} \manL(\zz) \approx \frac{1}{M} \sum^M_{i=1} \manL(I_\phi(w_i))
\end{equation}
by sampling $M$ realizations $w_1,\cdots,w_M$ from $p_w$.
The second term, however, is more difficult to evaluate since we lack the unknown and intractable $q_\phi$.
The second term is also called the (negative) entropy of $q_\phi$, usually denoted $H(q_\phi)\coloneqq -\EE_{\zz\sim q_\phi} \log q_\phi(\zz)$.
Fortunately, there are sample estimators $\hat H$ for $H$, so we can estimate it from a sample
% $z_1=I_\phi(w_1),\cdots,z_M=I_\phi(w_M)$.
$\{z_1, \cdots, z_M\},\; z_i=I_\phi(w_i)$.
We use the Kozachenko-Leonenko estimator~\citep{kozachenko1987sample,goria2005new},

\begin{equation}\label{eq:kozachenko}
  \hat H (\{z_i,\cdots,z_M\}) = \frac{n_z}{M} \sum^M_{i=1} \log \rho(z_i) + \mathrm{const.}
\end{equation}
where $\rho(z_i)$ is the distance between $z_i$ and its $k^{\mathrm{th}}$ nearest neighbor in the sample.
A good rule of thumb is $k\approx \sqrt{M}$~\citep{goria2005new}.
Intuitively, the entropy estimator measures how spread the elements of the sample are.
If the entropy term were not present, minimizing~\Cref{eq:kl} would reduce to finding the maximum a posteriori estimate, instead of sampling the full posterior.

To train the inference network $I_\phi$, we minimize $\operatorname{D_{KL}}(q_\phi \parallel p(\cdot|d_{\mathrm{obs}}))$ (\Cref{eq:kl}) using automatic differentiation algorithms.
% where both the estimator and the expected loss can be differentiated with respect to $\phi$ using automatic differentiation algorithms.
%
% Once the inference network is trained, the conditional generator is the new neural network $G\circ I\colon \RR^{n_w}\to\RR^{n_y}$, i.e. the composition of the unconditional generator and the inference network, as shown in~\Cref{fig:methodology}.
Once trained, we obtain our conditional parametrization $G_{\mathrm{cond}}=G\circ I\colon \RR^{n_w}\to\RR^{n_y}$.
Note that we now map from a new source distribution $\ww\sim p_w$, although we can simply pick $p_w = p_z$.
Sampling conditional realizations is done very efficiently by directly sampling $\ww\sim p_w$ and forward-passing through $G\circ I$.

We summarize the training steps of the inference network in~\Cref{algo:algo}. Note that we show a simple gradient descent update (line 7), however it is more common to use dedicated schemes for neural networks such as Adam~\citep{kingma2014adam} or RMSProp~\citep{tieleman2012lecture}.

Note that the inference network $I$ is relatively easy to train compared to the generator $G$ which is based on GAN. The network $I$ is also usually small and the relative increase in evaluation cost of the composition $G\circ I$ is not significant. We find this to be the case in our experiments.

\begin{algorithm}
  \caption{Inference network $I_\phi$ training}\label{algo:algo}
  \begin{algorithmic}[1]
  \Require{
    Negative log-posterior $\manL(z)=-\log p(z|d_{\mathrm{obs}})$. In our case (\Cref{eq:lossfn}),
    $\manL(z) = \|G(z)_{\mathrm{obs}} - d_{\mathrm{obs}}\|^2 +
    \lambda\|z\|^2$, batch size $M$, learning rate $\eta$, source distribution $p_w$ (usually equal to $p_z$).
    }
  %% \Ensure{$\theta$, $\psi$}
    \While{$\phi$ has not converged}

      \State Sample $\{w_1,\cdots,w_M\} \sim p_w$
      \State Get $\{z_1,\cdots,z_M\},\; z_i = I_\phi(w_i)$
      \State Get $\{\rho_1,\cdots,\rho_M\},\; \rho_i=$ distance from $z_i$ to its $k^{\mathrm{th}}$ nearest neighbor
      % \State $\nabla_\phi {\EE_{\zz\sim q_\phi} \manL(\zz)} \gets \frac{1}{M} \sum^M_{i=1} \nabla_\phi \manL(z_i)$
      \State $\nabla_\phi {\mathbb{E} \manL} \gets \frac{1}{M} \sum^M_{i=1} \nabla_\phi \manL(z_i)$
      \State $\nabla_\phi {\hat{H}} \gets \frac{n_z}{M} \sum^M_{i=1} \nabla_\phi {\log \rho_i}$
      \State $\phi \gets \phi - \eta (\nabla_\phi{\mathbb{E} \manL} - \nabla_\phi{\hat{H}})$

    \EndWhile
\end{algorithmic}
\end{algorithm}

% It is cumbersome to derive the complexity of deep learning techniques due to the large number of factors that come into play. The forward evaluation costs depend on the architecture, while the training cost depend on both architecture and ``difficulty of the problem''.
% Evaluation cost:
% The main issue is that the architecture can be made as small or large as required.
% For example, classifier type neural networks can take the form of simple linear classifiers $\operatorname{sign}{(w^Tx+b)}$, or can  1000-layers neural networks~\citep{kim2016deep}.
% $\manO(\alpha)$ where $\alpha$ is a number that conveys the size of the neural network architecture.
% Let $\alpha$ denote the number of operations required in an evaluation of the neural network.
% The issue is that $\alpha$ depends entirely on the architecture design of the neural network.
% For what is relevant in practice, we can assume $\alpha = akd$.
% Training cost:
% For training using backpropagation, $O(Tad)$.

% When $n\ll d$, as is often the case in geology, PCA $O(nd^2)$

% Our discussion highlight the difficulty in drawing a comparison of the complexities between deep learning techniques with other methods.
% Ultimately, the justification of the cost will largely depend on the particular problem that is addressed. Deep learning techniques are certainly justified in the traditional domains that they are developed, where deployment times are indefinitely long, e.g. recommender systems, ranking, etc. A similar case can be argued for subsurface modeling.

\section{Numerical experiments} \label{sec:results}

We train generative adversarial networks to obtain a parametrization of binary channelized subsurface images based on the benchmark image of~\citet{strebelle2001reservoir}.
We then condition the parametrization for a variety of configurations using our method described in~\Cref{sec:methodology}.
Finally, we also include in~\Cref{sec:mixture} a side experiment as a sanity check of the proposed method where we train neural samplers for simple mixture of Gaussians.
All our numerical experiments are implemented using PyTorch~\citep{paszke2017automatic}, an open-source Python package for automatic differentiation that provides tools to facilitate the construction and training of neural network models.
Our implementation code is available in our repository\footnote{\url{https://github.com/chanshing/geocondition}}.

\subsection{Unconditional parametrization}\label{sec:results_uncond}

\begin{figure}\centering

\begin{figure}[H]\centering

    \begin{subfigure}{\textwidth}\centering
        \includegraphics[width=.9\textwidth]{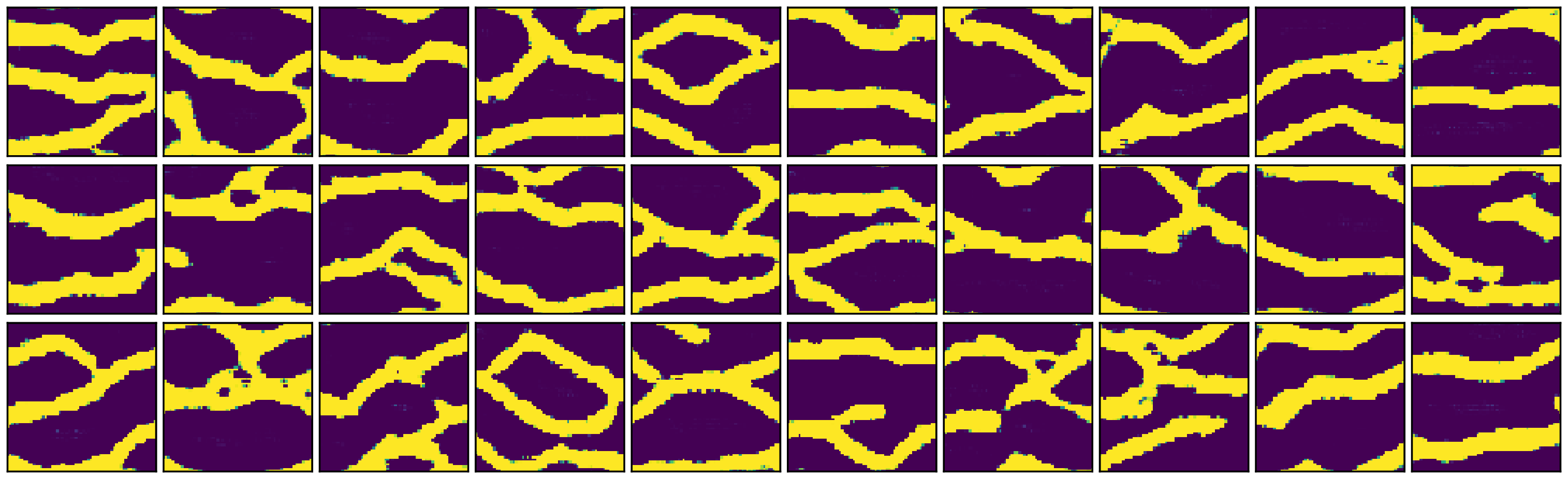}
        \caption{$G(\zz)$\label{fig:gen_uncond}}
    \end{subfigure}\vspace{1em}

    \begin{subfigure}{\textwidth}\centering
        \includegraphics[width=.9\textwidth]{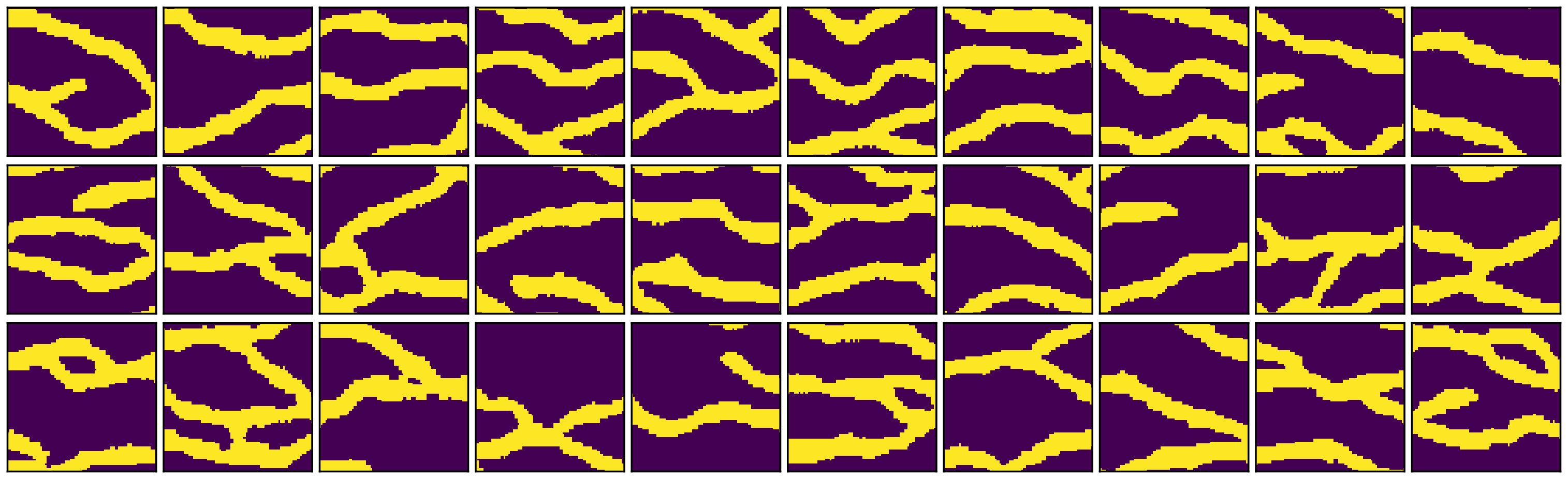}
        \caption{\texttt{snesim}\label{fig:snesim_uncond}}
    \end{subfigure}

    \begin{subfigure}{\textwidth}\centering
        \includegraphics[width=\textwidth]{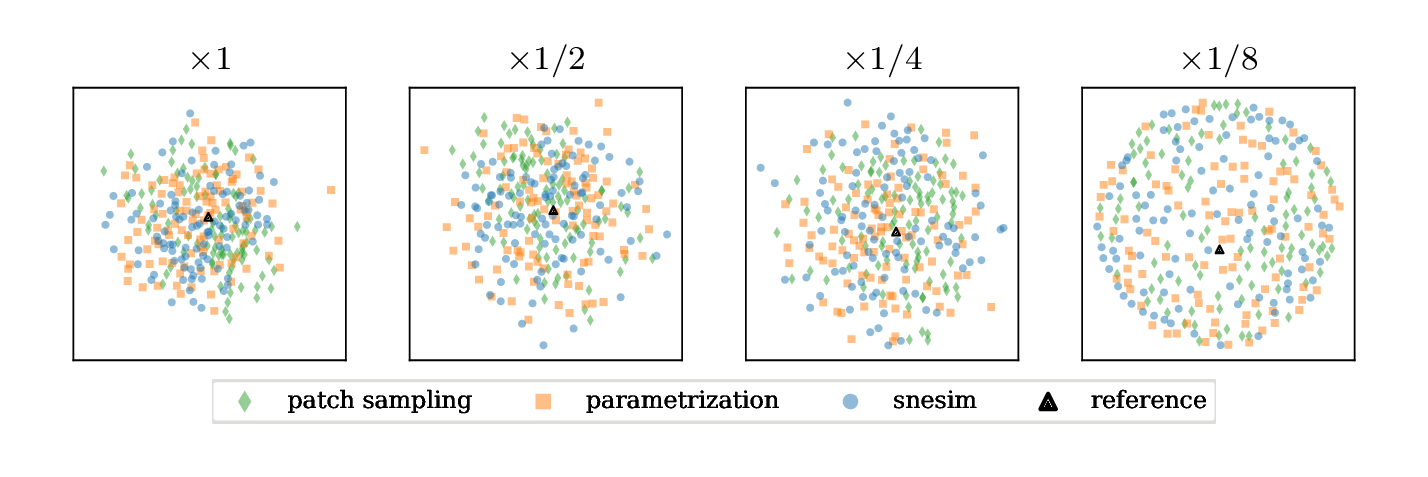}
        \vspace{-2em}\caption{Multidimensional scaling visualization.\label{fig:mds_uncond}}
    \end{subfigure}

    \caption{Unconditional realizations.\label{fig:uncond}}

\end{figure}

\vspace{1em}
\begin{table}[H]\centering
    \caption{Unconditional realizations. ANODI scores (inconsistency/diversity).\label{table:uncond}}
    \vspace{-1.5em}
    \includegraphics[width=\textwidth]{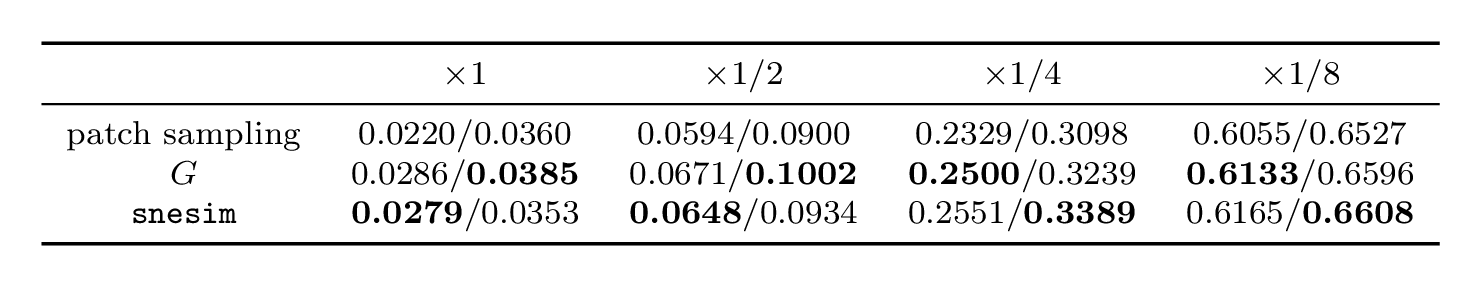}
\end{table}

\end{figure}

We train a generator $G\colon\RR^{30}\to\RR^{64\times64}$ using a dataset of $1000$ realizations of size $64\times 64$ of binary channelized subsurface images.
The realizations were obtained using the \texttt{snesim} algorithm~\citep{strebelle2001reservoir,strebelle2002conditional} provided within the Stanford geostatistical modeling software~\citep{remy2004sgems}, using the benchmark image from~\citet{strebelle2001reservoir} as the reference image\footnote{Also referred to as a \emph{training image} in the geostatistics literature, although we avoid the term so that it is not confused with the images of the training set used to train the neural network.}. A few \texttt{snesim} realizations are shown in~\Cref{fig:snesim_uncond}.

\subsubsection{Architecture design}
The latent vector size $n_z=30$ was chosen using principal component analysis as a heuristic, where the number of eigencomponents required to retain 75\% of the variance is used as a reference. This results in a dimensionality reduction of two orders of magnitude -- from $64\times 64$ to 30.
The latent vector is sampled from the standard normal distribution, $\zz\sim\mathcal{N}(\mathbf{0},\mathbf{I})$.
Since the data is binary, it is reasonable to embed this knowledge into the neural network design using a suitable non-linearity in the output layer of the neural network. We use $\sigma = \tanh$, where we adopt $1$ to denote channel material, and $-1$ to denote background material.
Note that attempting to use a hard threshold here would render $G$ discontinuous and introduce issues in the training. It can also be an issue in inversion problems during deployment.
The rest of the neural network architecture (shapes of $A_i, b_i$, non-linearity $\sigma_i$, number of layers $l$, etc. -- see~\Cref{eq:dnn}) is designed according to the template provided in~\citep{radford2015unsupervised}.
This template is the result of experimentation, heuristics, and experience. In particular, an important design choice is the use of convolutional layers. These are sparse matrices $A_i$ that follow a certain structure that makes them effective for spatial data. A brief description of convolutional layers is provided in~\Cref{sec:cnn}.
Further details of the generator architecture and training is given in~\Cref{sec:details_gen}.

\subsubsection{Quality assessment}

Realizations generated by the parametrization $G$ are shown in~\Cref{fig:gen_uncond}.
We also show \texttt{snesim} realizations in~\Cref{fig:snesim_uncond}.
We can already see from the figure that the parametrization is at least visually competitive with previous parametrization methods.
The realizations of the parametrization are virtually indistinguishable from \texttt{snesim},
recreating crisp and clear channels from the reference (note that no thresholding has been performed).
We next assess the results quantitatively.

Previous works have assessed the effectiveness of GAN-based parametrization using a variety of tools. Two-point probability functions, morphological measures, and effective porosity were assessed in~\citep{mosser2017reconstruction}. Two-point probability and cluster functions, and fractions of facies were assessed in~\citep{laloy2018training}.
In our previous work~\citep{chan2017parametrization}, we assessed the effectiveness of the parametrization for preserving high-order flow statistics in uncertainty quantification.
Here we add to the assessment using the method of analysis of distances (ANODI)~\citep{tan2014comparing} which captures multipoint statistics, providing a more reliable measure of quality for complex data where two-point statistics are insufficient. We also apply multidimensional scaling for visualization.

The ANODI method aims to capture multipoint statistics by comparing multipoint histograms at different resolutions. It computes an inconsistency score (how well it matches the statistics of the reference image) and a diversity score (variability between realizations) -- therefore, we want low inconsistency and high diversity.
Multidimensional scaling is a method that aims to project a set of high-dimensional objects to low dimensions in a way that preserves the distances between the objects. Although some information may be lost in the projection, the method provides a useful way of visualizing high-dimensional objects (e.g. images) using a scatter-plot.
The notion of distance between images, as adopted in~\citep{tan2014comparing}, is the Jensen-Shannon divergence between multipoint histograms of patterns extracted from the images within a window size.

We perform the analysis at four resolutions: $\times 1$ (original), $\times 1/2$, $\times 1/4$, and $\times 1/8$ resolution (i.e. at $64\times 64$, $32\times 32$, $16\times 16$, and $8\times 8$). We use a window size of $4\times 4$ and sets of 100 realizations for the analysis. Importantly, note that the \texttt{snesim} realizations are fresh realizations, i.e not from the training dataset: Ultimately, we aim to plug the parametrization into a reservoir simulator, for which we are assuming that the parametrization replicates the data generating process. Therefore, the comparison is made against out-of-sample realizations to see if the parametrization generalizes.
For multidimensional dimensional scaling, we use the SMACOF~\citep{borg2003modern} algorithm with 300 iterations and tolerance of $10^{-3}$.

For the analysis, we need to binarize the realizations generated by the parametrization which are continuous by design. For this, we use Otsu's thresholding method~\citep{otsu1979threshold}. We also apply small object removal processing on the images to remove possible isolated pixels. Note that we do \emph{not} apply thresholding nor any other post-processing on the displayed images in this work.
Finally, since it can be difficult to gauge the differences in the ANODI scores, we include ``patch sampling'' method (i.e. drawing patches of $64\times 64$ from the reference image) into the analysis to serve as a third point of comparison.

We show the ANODI scores and multidimensional scaling visualizations in~\Cref{table:uncond} and~\Cref{fig:mds_uncond}, respectively.
The patch sampling procedure understandably produces the highest consistency (lowest inconsistency), although it is also slightly less diverse.
Regarding the parametrization, we find that the scores for \texttt{snesim} and $G$ are relatively very close across all resolutions, suggesting that $G$ effectively learned to replicate the data generating process.
The multidimensional scaling visualization in~\Cref{fig:mds_uncond} further supports this result, showing a very good overlap in the scatter-plots of \texttt{snesim} and $G$. The scatter-plots are also well spread and centered around the reference image, verifying the good performance of both methods.

\subsubsection{Memorization}\label{sec:memorization}

\begin{figure}\centering
    \begin{subfigure}{\textwidth}\centering
    \includegraphics[width=\textwidth]{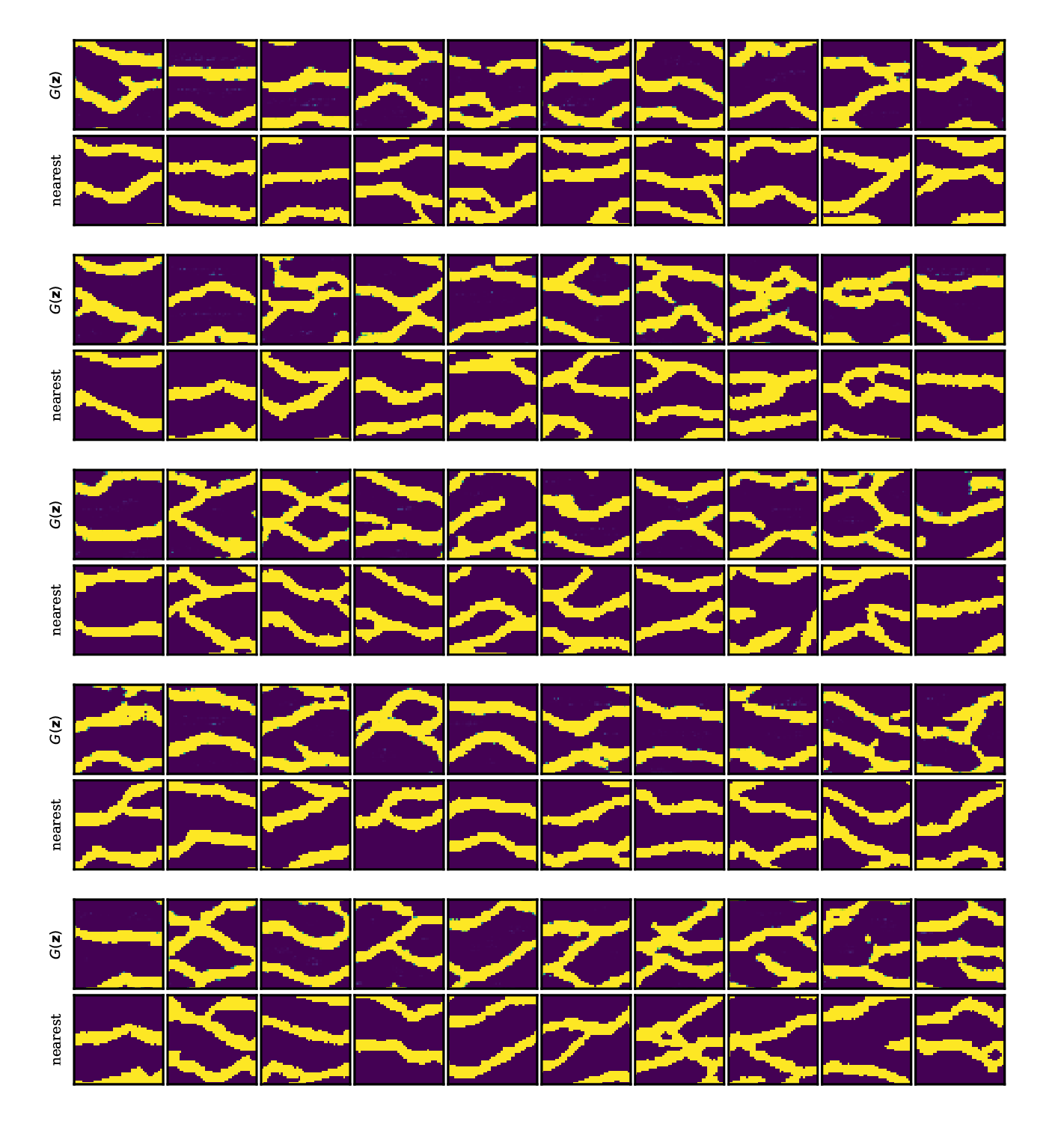}
    \vspace{-2.5em}
    \caption{Nearest neighbors for 50 generated realizations.\label{fig:nearest}}
    \end{subfigure}\vspace{2em}

    \begin{subfigure}{\textwidth}\centering
    \includegraphics[width=.955\textwidth]{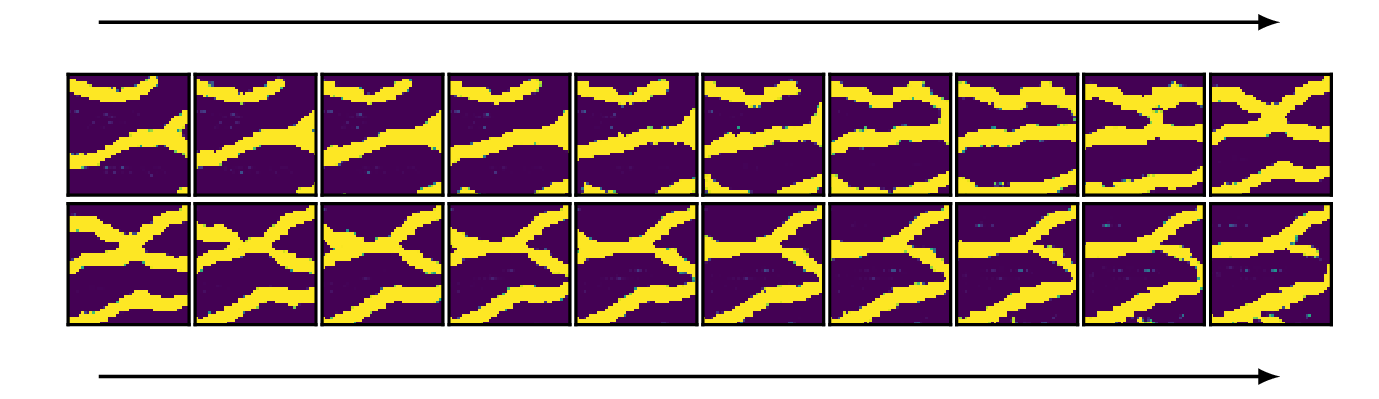}
    \caption{Interpolation in the latent space.\label{fig:interpol}}
    \end{subfigure}
    \caption{Assessment of memorization of $G$.}
\end{figure}

To verify that the parametrization is not simply memorizing the training dataset, we find the nearest neighbor in the dataset for each of 50 generated realizations. To further verify that the generator is not simply learning trivial transformations, we data-augment our training dataset with horizontal and vertical flips, as well as $10$ and $-10$ degrees rotation and shearing (with reflection filling at the boundaries). This results in 35 additional variations for each image in the dataset. Finally, to capture small translations, we apply a Gaussian blur to the images before computing the Euclidean distance.

The 50 realizations along with the nearest neighbors are shown in~\Cref{fig:nearest}.
We see that there is no perfect match despite the heavy data augmentation, verifying that the parametrization is capable of generating novel realizations that are not mere rotations, translations, shearing and flips of images from the dataset.
The lack of memorization can be justified by the fact that the generator never has direct access to the training dataset (see~\Cref{eq:ganloss}). Instead, the generator only obtains indirect information about the dataset via the discriminator (i.e. through its gradients).
This is similar to principal component analysis where the parametrization is only informed about the dataset covariance. In the case of GAN, the relevant dataset statistics are automatically discovered and informed by the discriminator.

Finally, we provide a further verification by performing an interpolation in the latent space in~\Cref{fig:interpol}.
If $G$ is simply memorizing the dataset, we would expect to see sudden jumps from one image of the dataset to another, with implausible images in between as we interpolate in the latent space. We instead effectively find smooth transitions between plausible outputs.
The smoothness is also justified by the fact that $G$ is continuous and piecewise differentiable by design (see~\Cref{eq:dnn}).
Note also that the smoothness is critical in practice for efficient exploration of the solution space during deployment, e.g. for inversion and uncertainty quantification tasks.

\subsection{Conditional parametrization}

\begin{figure}\centering

    \begin{figure}[H]\centering
        \begin{subfigure}{\textwidth}\centering
            \includegraphics[width=.9\textwidth]{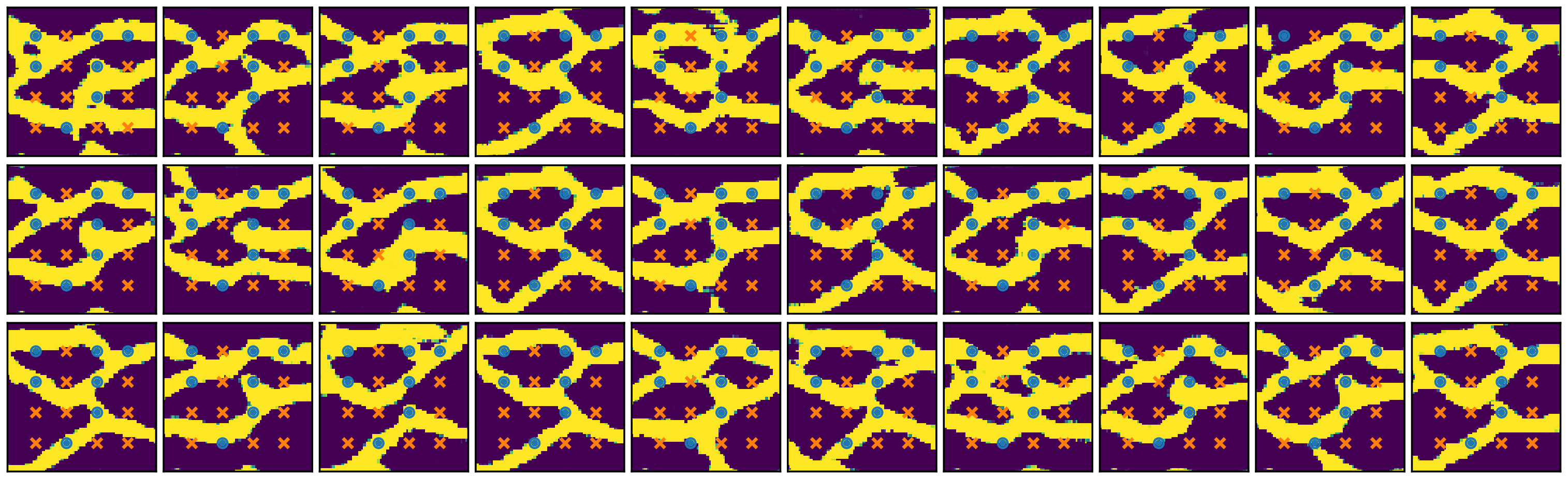}
            \caption{$G\circ I(\ww)$ \label{fig:gen_cond01}}
        \end{subfigure}\vspace{1em}

        \begin{subfigure}{\textwidth}\centering
            \includegraphics[width=.9\textwidth]{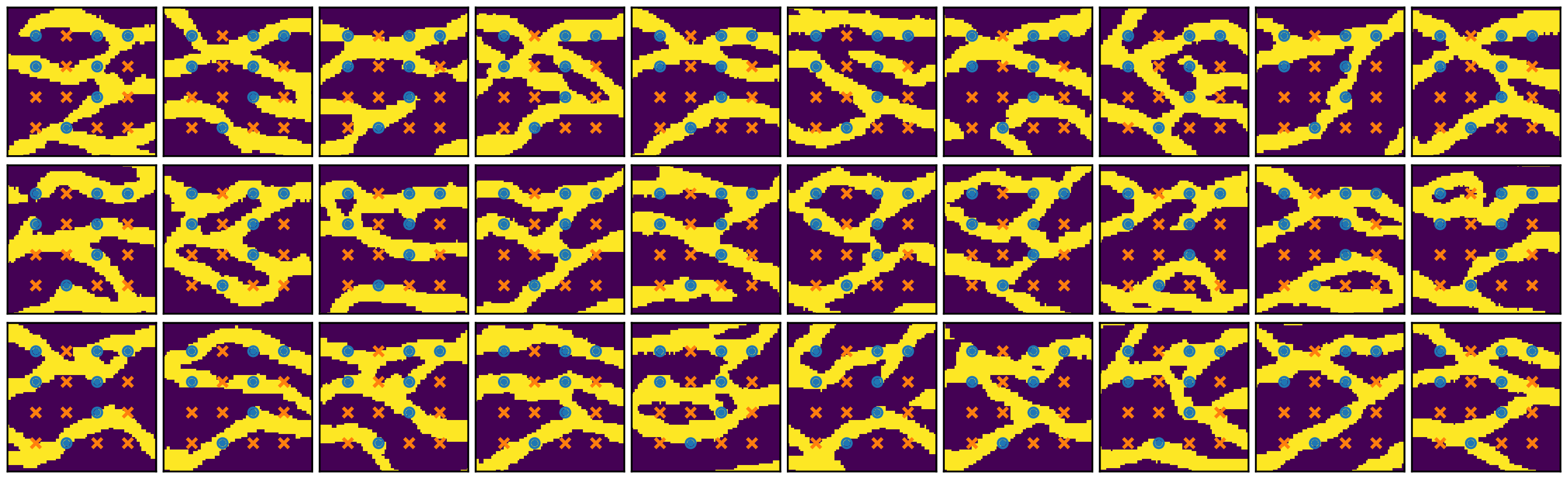}
            \caption{\texttt{snesim} \label{fig:snesim_cond01}}
        \end{subfigure}

        \begin{subfigure}{\textwidth}\centering
            \includegraphics[width=\textwidth]{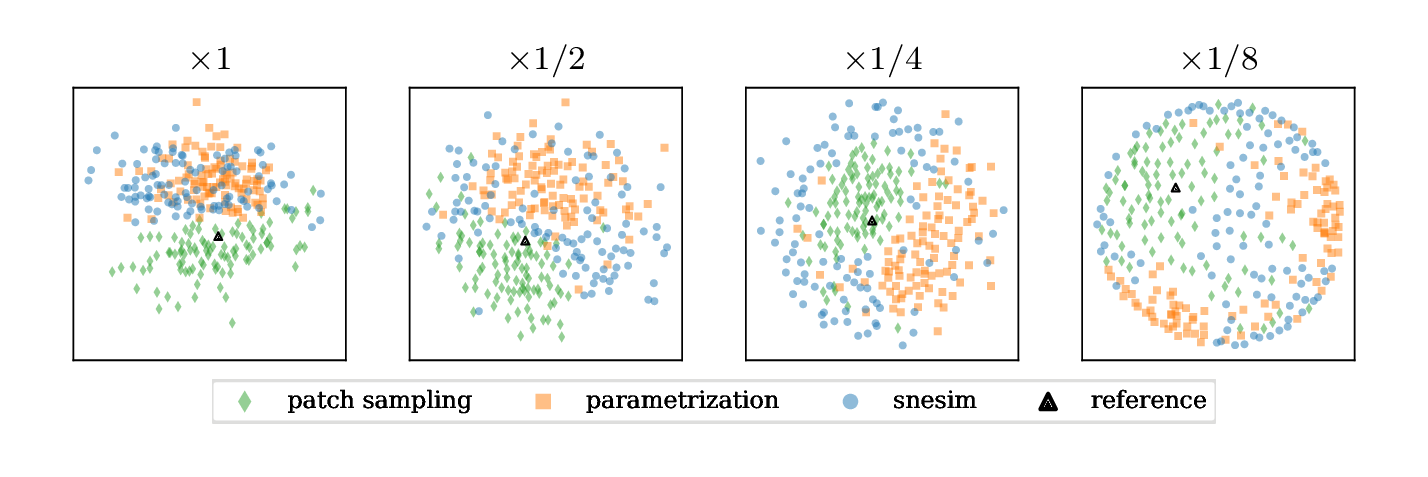}
            \vspace{-2em}\caption{Multidimensional scaling visualization.\label{fig:mds_cond01}}
        \end{subfigure}

        \caption{Example A. Conditional realizations.\label{fig:cond01}}

    \end{figure}

    \vspace{1em}
    \begin{table}[H]\centering
        \caption{Example A. ANODI scores (inconsistency/diversity).\label{table:cond01}}
        \vspace{-1.5em}
        \includegraphics[width=\textwidth]{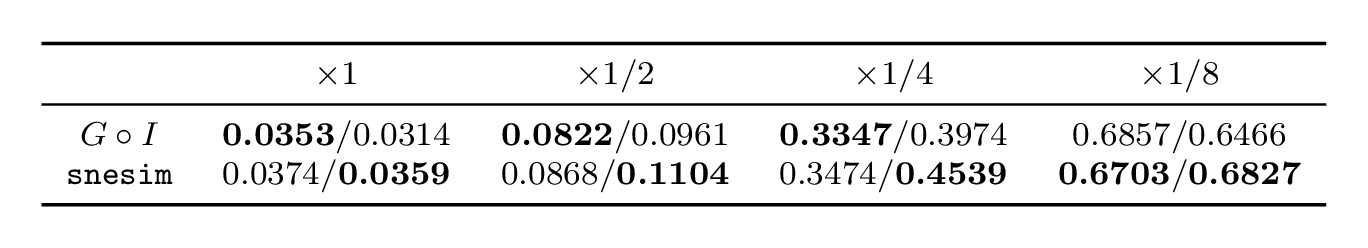}
    \end{table}

\end{figure}

\begin{figure}\centering

    \begin{figure}[H]\centering
        \begin{subfigure}{\textwidth}\centering
            \includegraphics[width=.9\textwidth]{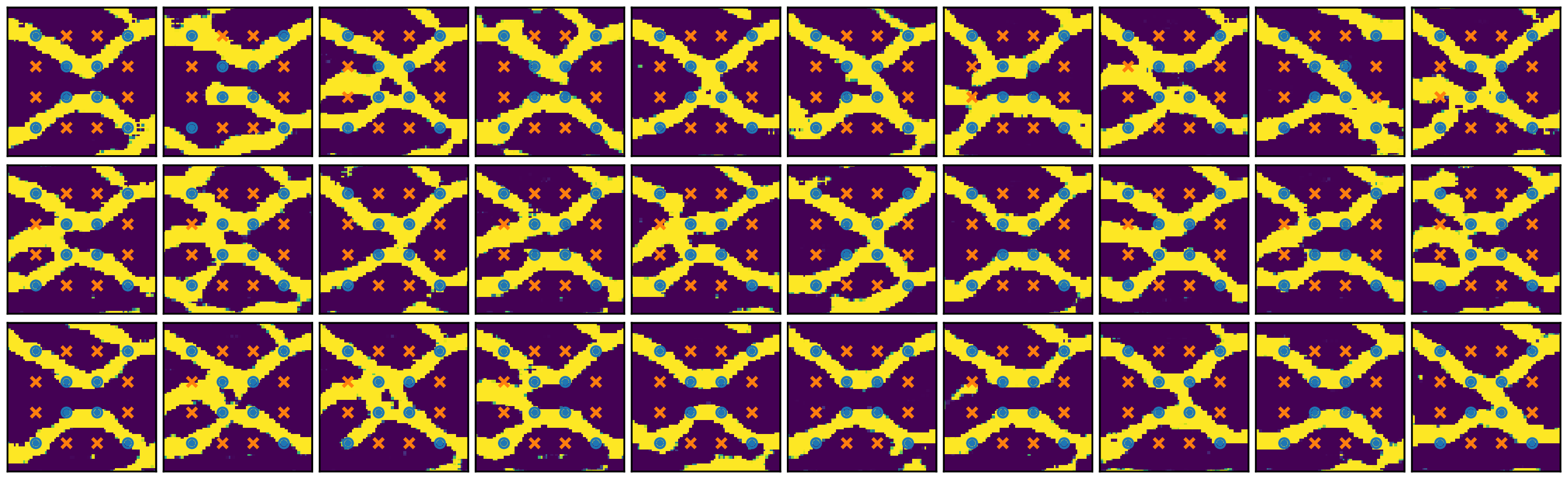}
            \caption{$G\circ I(\ww)$\label{fig:gen_cond02}}
        \end{subfigure}\vspace{1em}

        \begin{subfigure}{\textwidth}\centering
            \includegraphics[width=.9\textwidth]{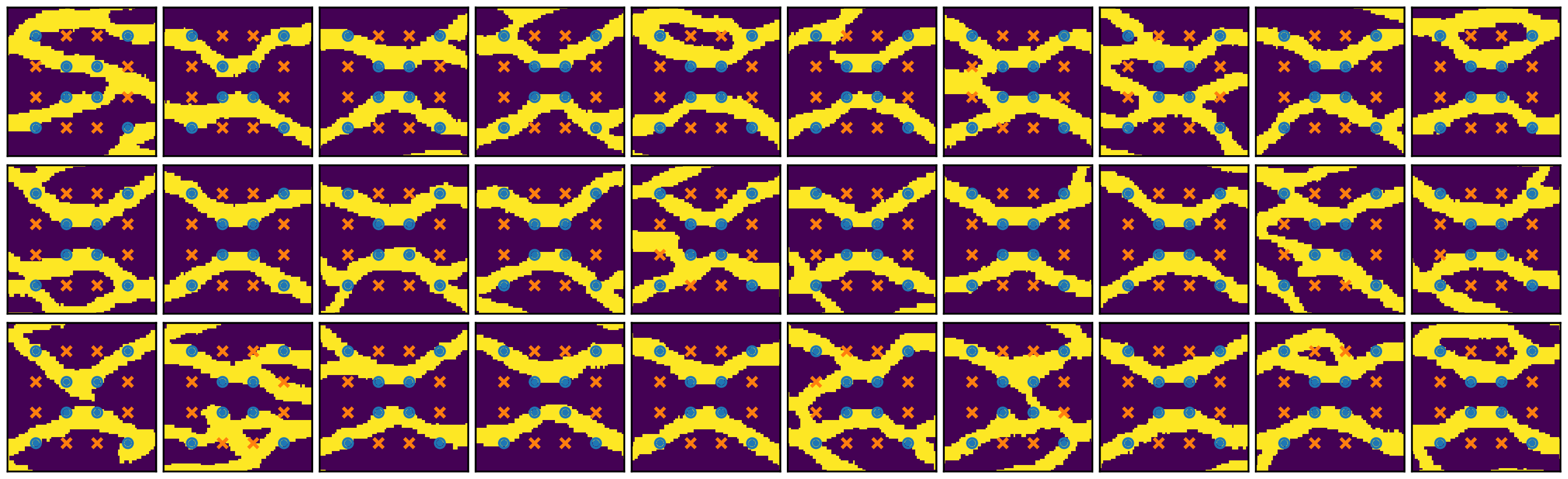}
            \caption{\texttt{snesim}\label{fig:snesim_cond02}}
        \end{subfigure}

        \begin{subfigure}{\textwidth}\centering
            \includegraphics[width=\textwidth]{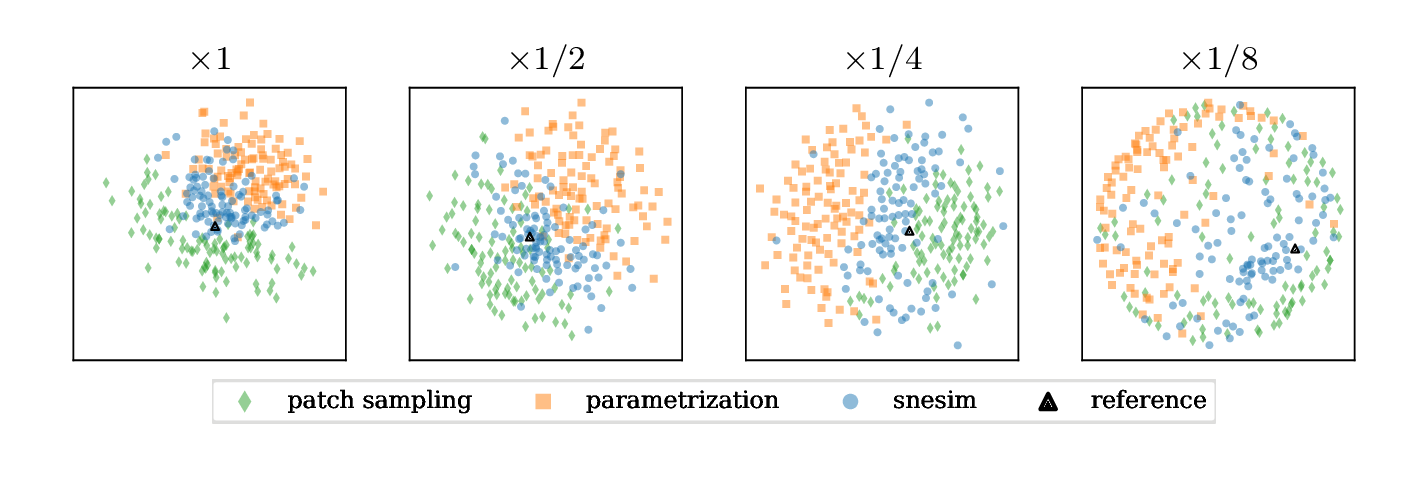}
            \vspace{-2em}\caption{Multidimensional scaling visualization.\label{fig:mds_cond02}}
        \end{subfigure}

        \caption{Example B. Conditional realizations.\label{fig:cond02}}
    \end{figure}

        \vspace{1em}
        \begin{table}[H]\centering
            \caption{Example B. ANODI scores (inconsistency/diversity).\label{table:cond02}}
            \vspace{-1.5em}
            \includegraphics[width=\textwidth]{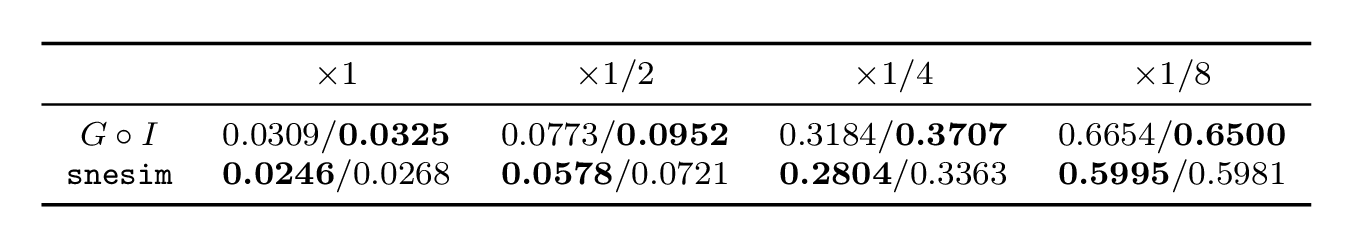}
        \end{table}

\end{figure}

\begin{figure}\centering
    \begin{figure}[H]\centering
        \begin{subfigure}{\textwidth}\centering
            \includegraphics[width=.9\textwidth]{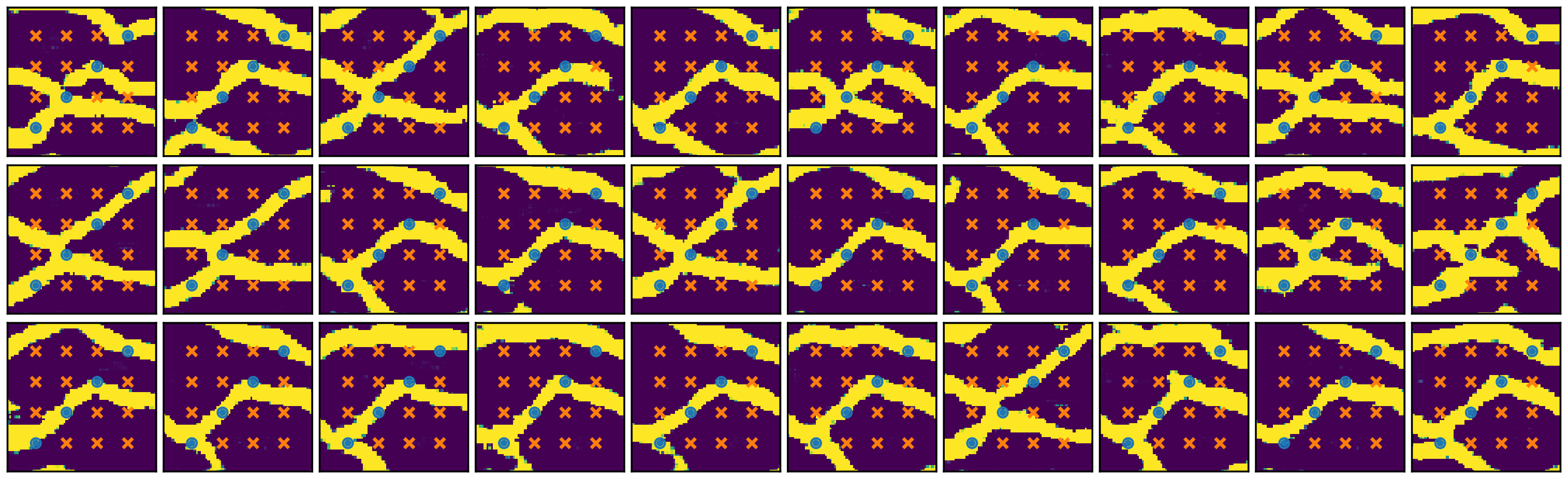}
            \caption{$G\circ I(\ww)$\label{fig:gen_cond03}}
        \end{subfigure}\vspace{1em}

        \begin{subfigure}{\textwidth}\centering
            \includegraphics[width=.9\textwidth]{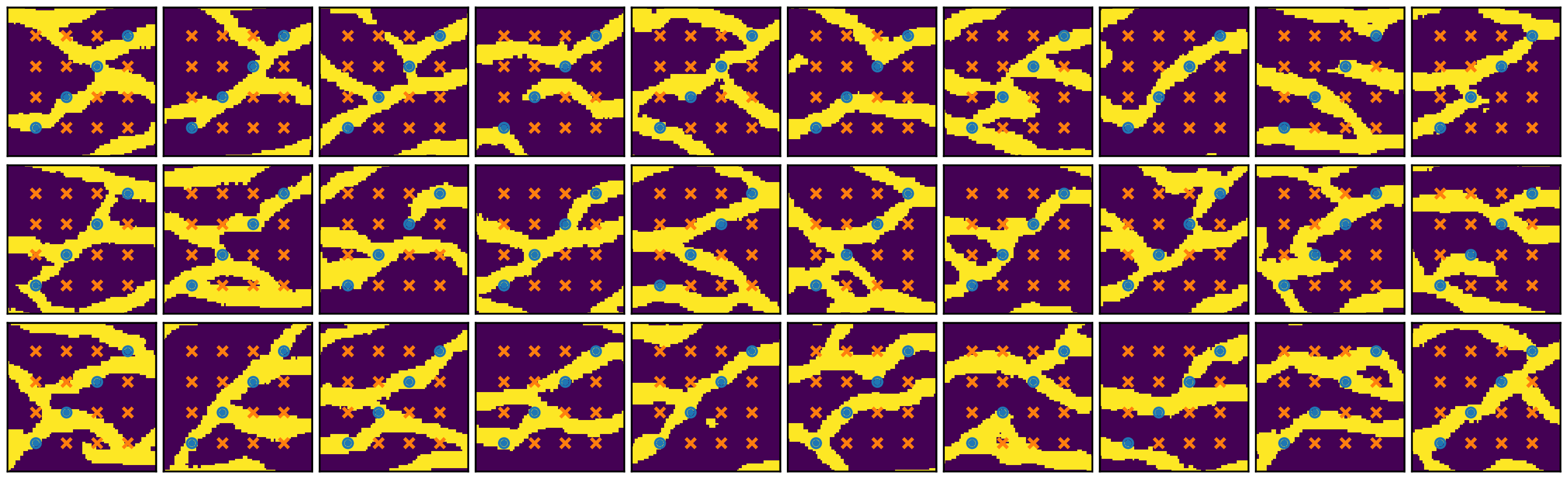}
            \caption{\texttt{snesim}\label{fig:snesim_cond03}}
        \end{subfigure}

        \begin{subfigure}{\textwidth}\centering
            \includegraphics[width=\textwidth]{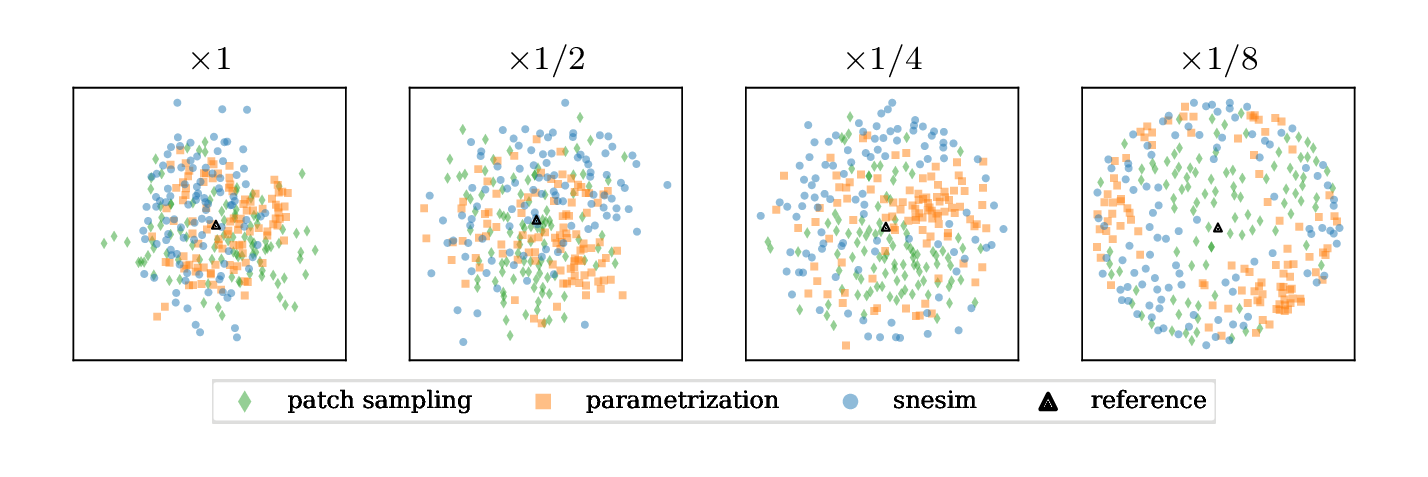}
            \vspace{-2em}\caption{Multidimensional scaling visualization.\label{fig:mds_cond03}}
        \end{subfigure}

        \caption{Example C. Conditional realizations.\label{fig:cond03}}
    \end{figure}

    \vspace{1em}
    \begin{table}[H]\centering
        \caption{Example C. ANODI scores (inconsistency/diversity).\label{table:cond03}}
        \vspace{-1.5em}
        \includegraphics[width=\textwidth]{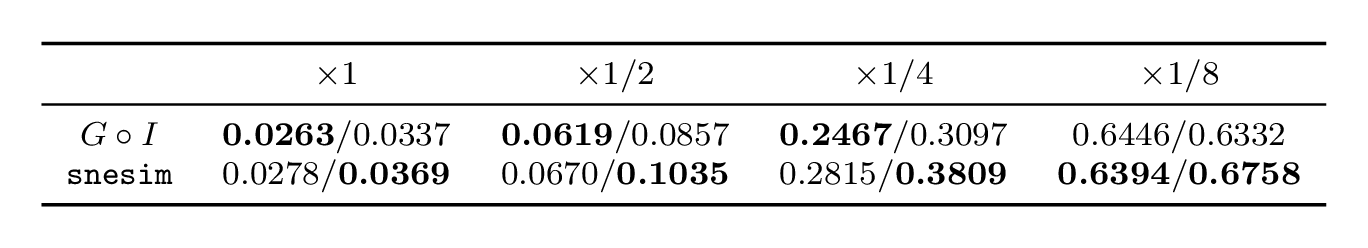}
    \end{table}

\end{figure}

\begin{figure}
	\begin{figure}[H]\centering
		\begin{subfigure}{\textwidth}\centering
			\includegraphics[width=.9\textwidth]{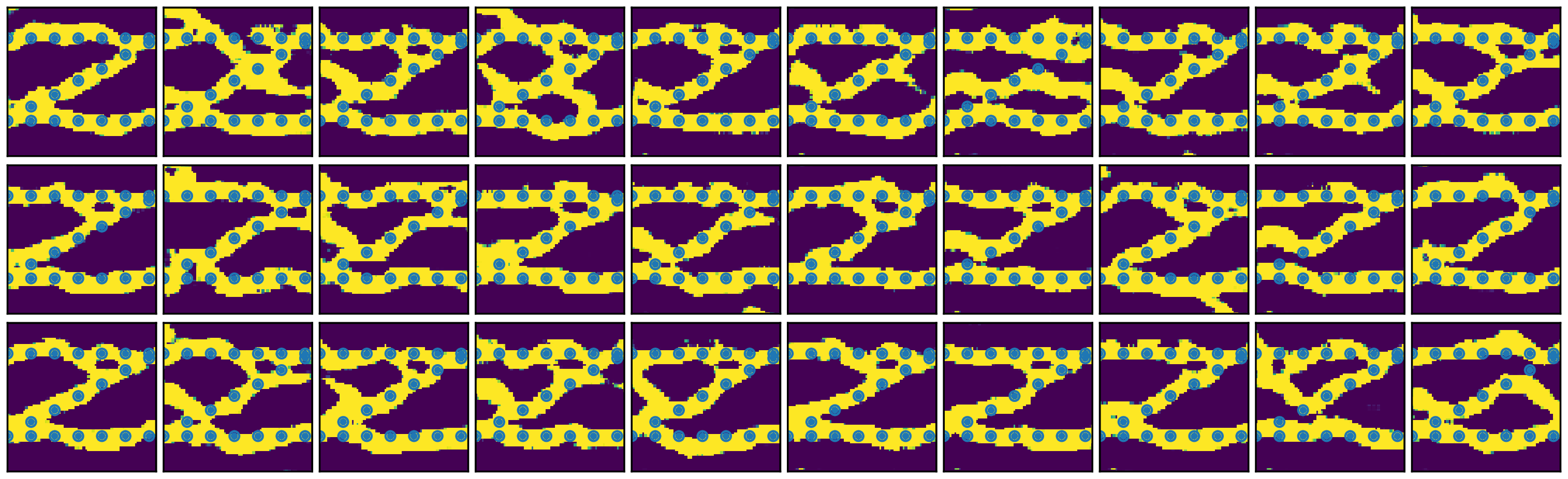}
			\caption{$G\circ I(\ww)$\label{fig:gen_cond11}}
		\end{subfigure}\vspace{1em}

		\begin{subfigure}{\textwidth}\centering
			\includegraphics[width=.9\textwidth]{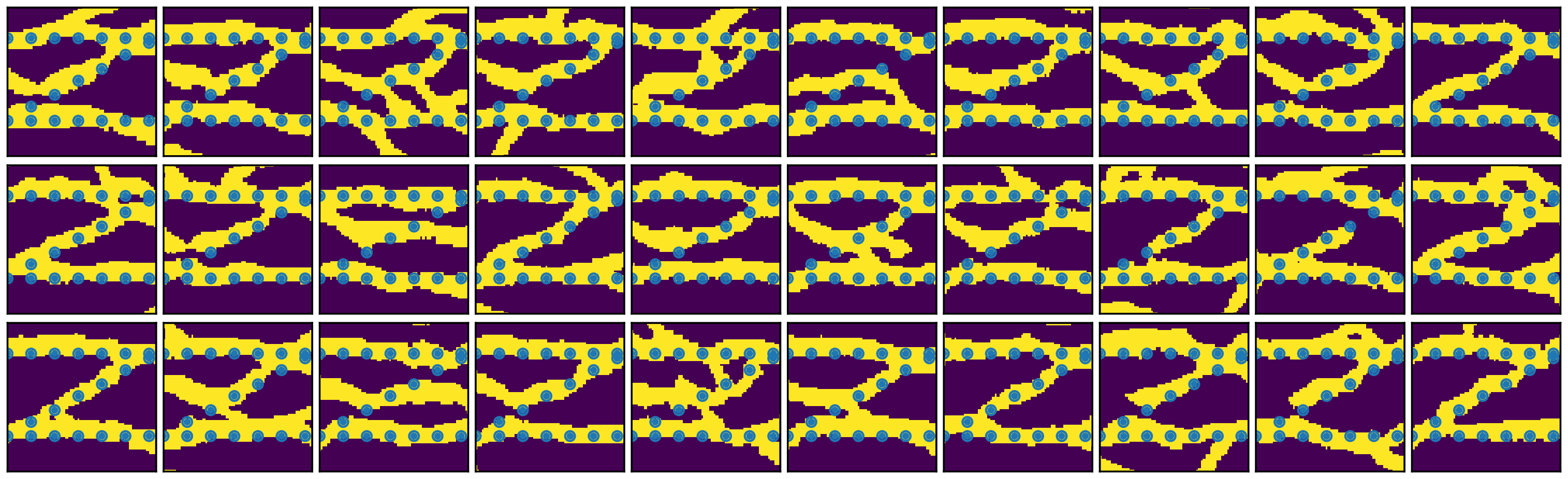}
			\caption{\texttt{snesim}\label{fig:snesim_cond11}}
		\end{subfigure}

		\begin{subfigure}{\textwidth}\centering
			\includegraphics[width=\textwidth]{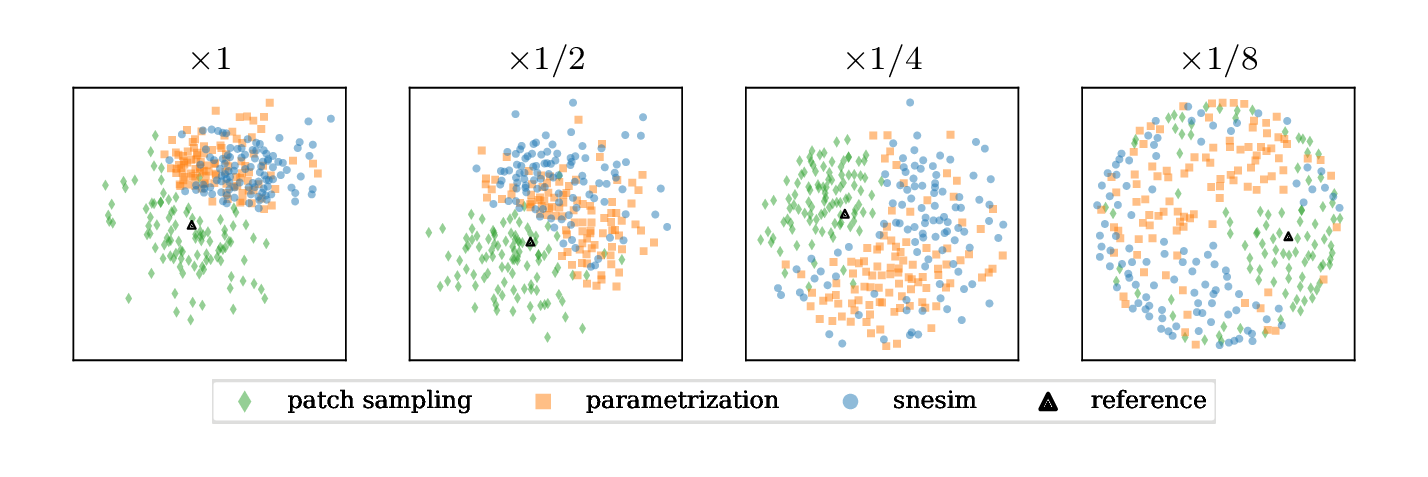}
			\vspace{-2em}\caption{Multidimensional scaling visualization.\label{fig:mds_cond11}}
		\end{subfigure}

		\caption{Example D. Conditional realizations.\label{fig:cond11}}
	\end{figure}

	\vspace{1em}
	\begin{table}[H]\centering
		\caption{Example D. ANODI scores (inconsistency/diversity).\label{table:cond11}}
		\vspace{-1.5em}
		\includegraphics[width=\textwidth]{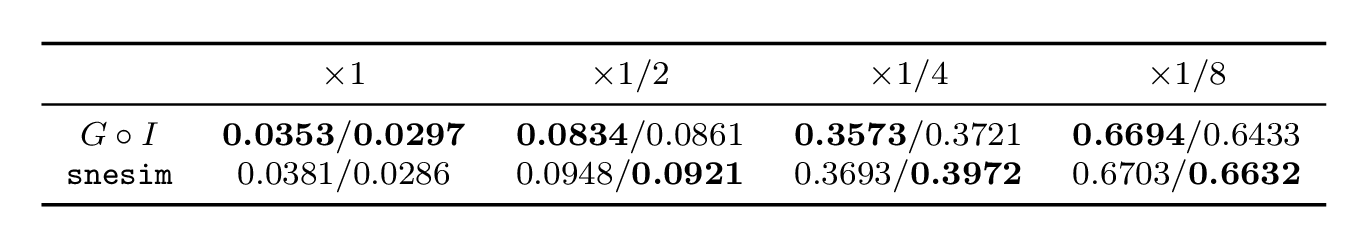}
	\end{table}
\end{figure}

\begin{figure}
	\begin{figure}[H]\centering
		\begin{subfigure}{\textwidth}\centering
			\includegraphics[width=.9\textwidth]{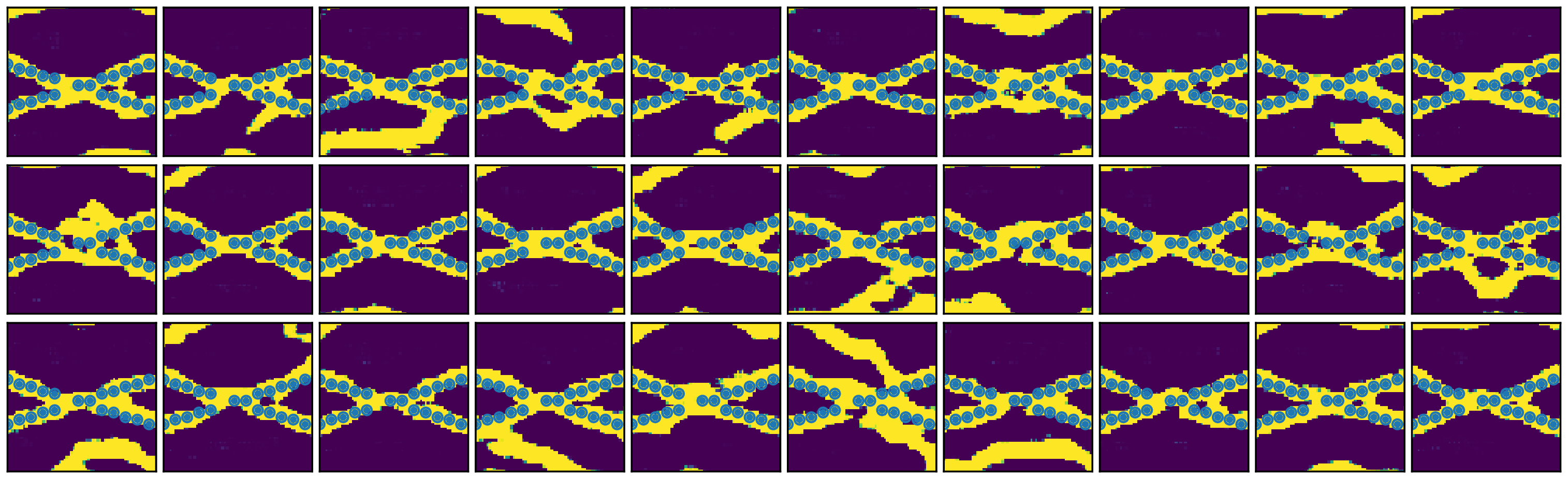}
			\caption{$G\circ I(\ww)$\label{fig:gen_cond12}}
		\end{subfigure}\vspace{1em}

		\begin{subfigure}{\textwidth}\centering
			\includegraphics[width=.9\textwidth]{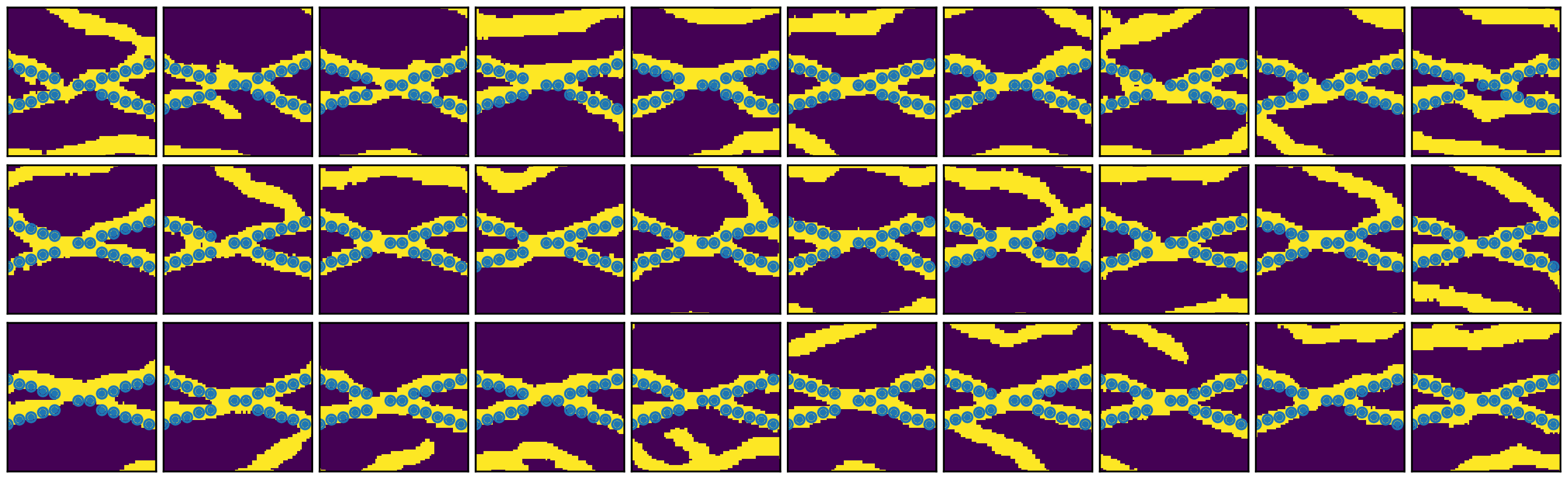}
			\caption{\texttt{snesim}\label{fig:snesim_cond12}}
		\end{subfigure}

		\begin{subfigure}{\textwidth}\centering
			\includegraphics[width=\textwidth]{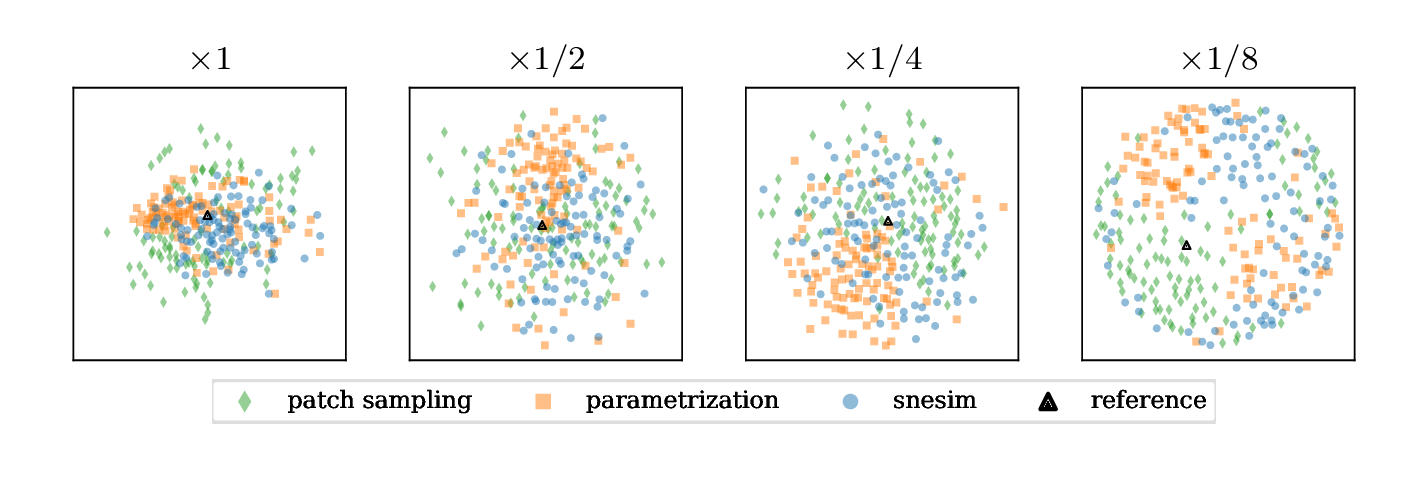}
			\vspace{-2em}\caption{Multidimensional scaling visualization.\label{fig:mds_cond12}}
		\end{subfigure}

		\caption{Example E. Conditional realizations.\label{fig:cond12}}
	\end{figure}

	\vspace{1em}
	\begin{table}[H]\centering
		\caption{Example E. ANODI scores (inconsistency/diversity).\label{table:cond12}}
		\vspace{-1.5em}
		\includegraphics[width=\textwidth]{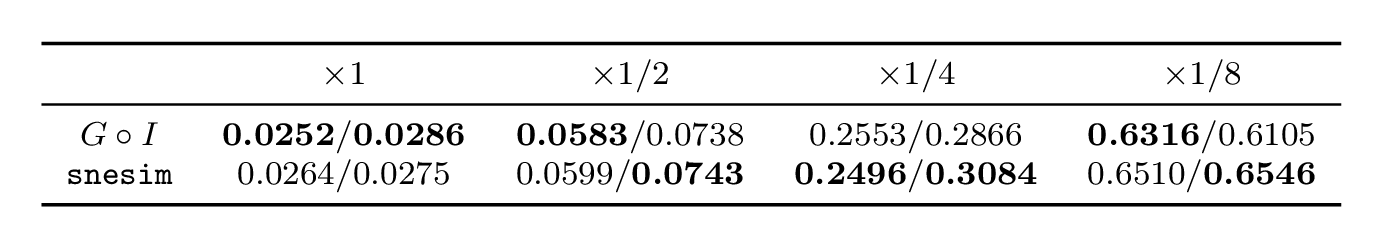}
	\end{table}
\end{figure}

\begin{figure}\centering
	\begin{figure}[H]\centering
		\begin{subfigure}{\textwidth}\centering
			\includegraphics[width=.9\textwidth]{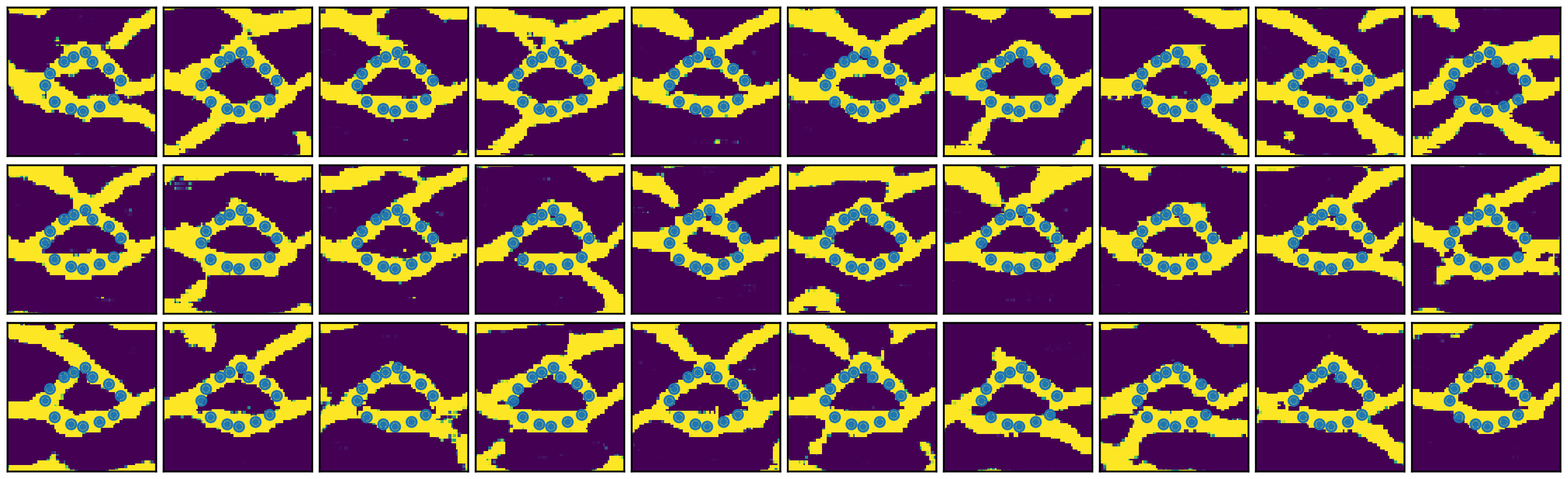}
			\caption{$G\circ I(\ww)$\label{fig:gen_cond13}}
		\end{subfigure}\vspace{1em}

		\begin{subfigure}{\textwidth}\centering
			\includegraphics[width=.9\textwidth]{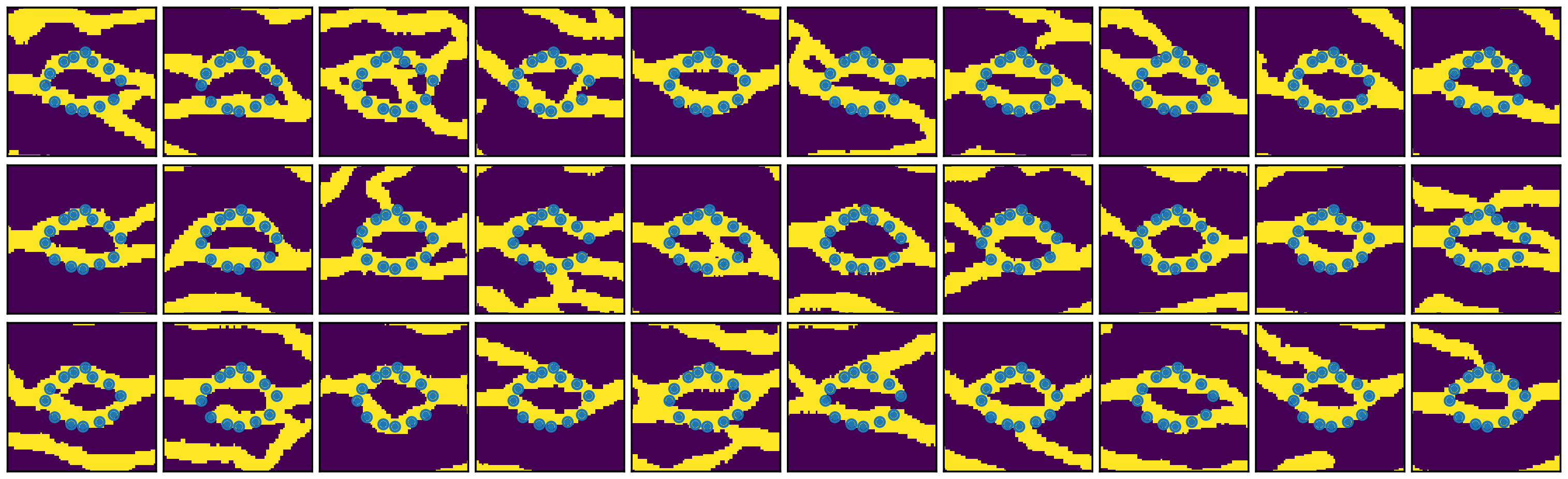}
			\caption{\texttt{snesim}\label{fig:snesim_cond13}}
		\end{subfigure}

		\begin{subfigure}{\textwidth}\centering
			\includegraphics[width=\textwidth]{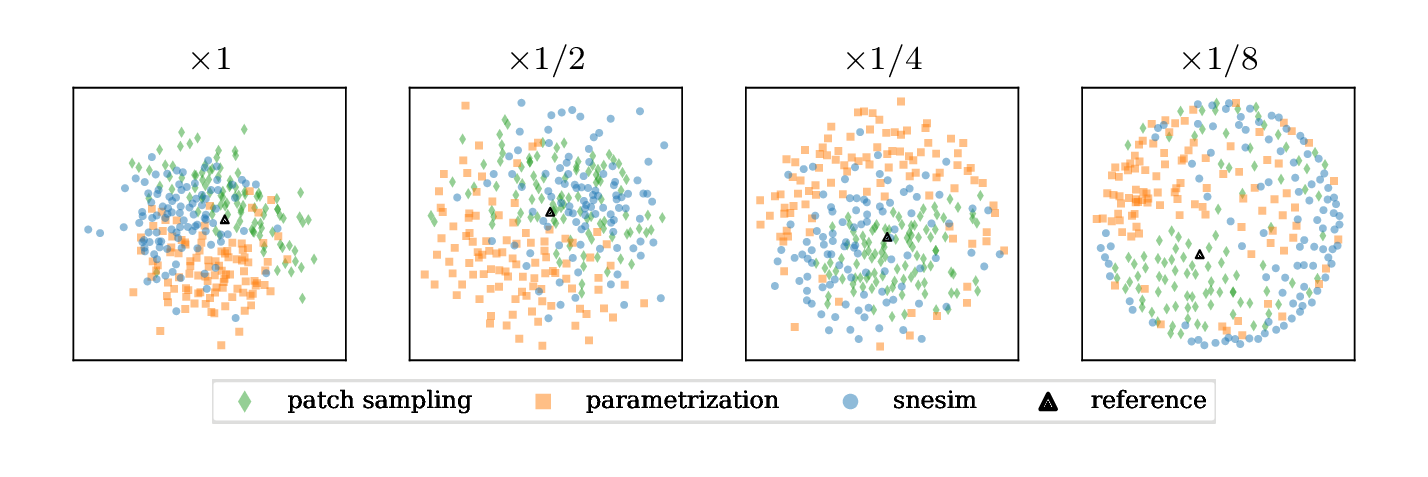}
			\vspace{-2em}\caption{Multidimensional scaling visualization.\label{fig:mds_cond13}}
		\end{subfigure}

		\caption{Example F. Conditional realizations.\label{fig:cond13}}
	\end{figure}

	\vspace{1em}
	\begin{table}[H]\centering
		\caption{Example F. ANODI scores (inconsistency/diversity).\label{table:cond13}}
		\includegraphics[width=\textwidth]{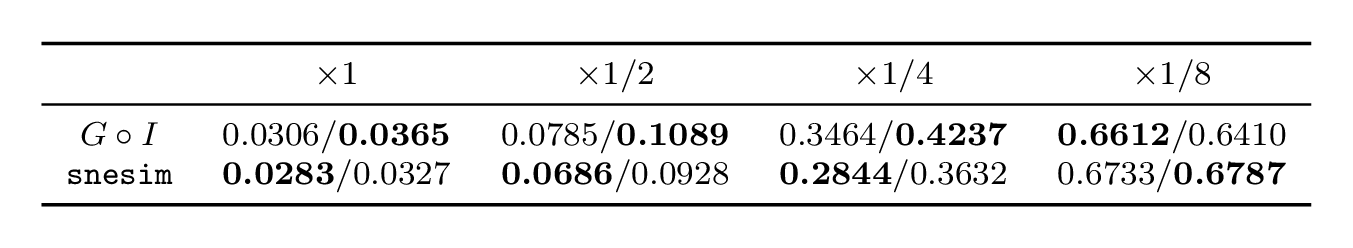}
		\vspace{-1.5em}
	\end{table}
\end{figure}

\begin{figure}\centering
	\begin{figure}[H]\centering
		\begin{subfigure}{\textwidth}\centering
			\includegraphics[width=.9\textwidth]{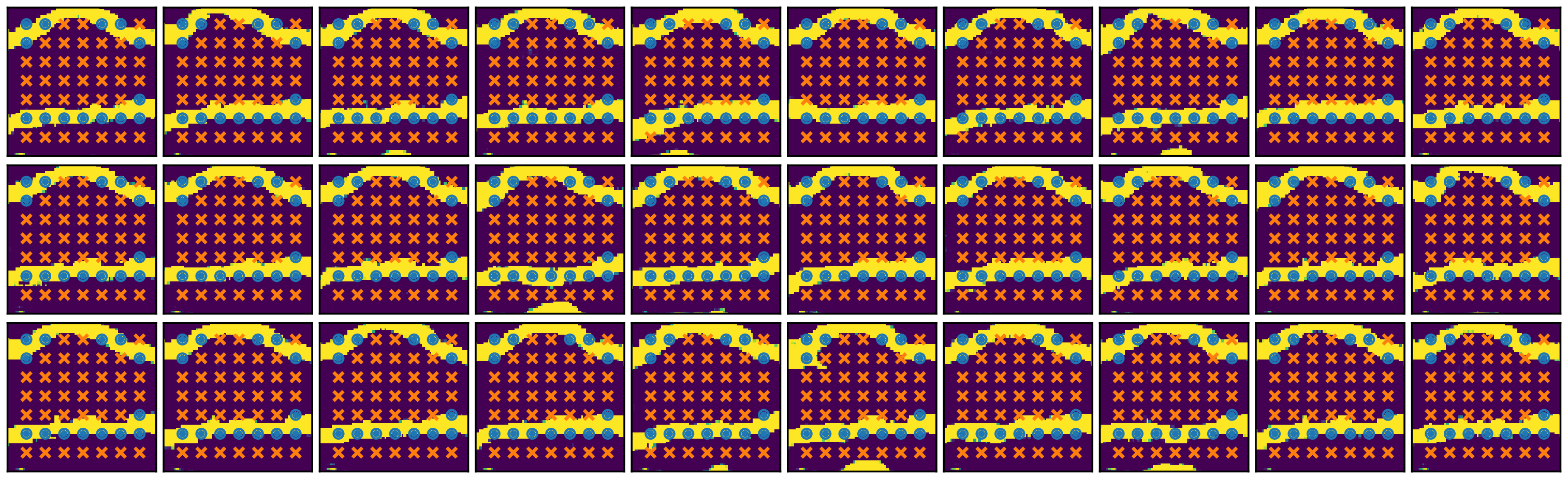}
			\caption{$G\circ I(\ww)$\label{fig:gen_cond21}}
		\end{subfigure}\vspace{1em}

		\begin{subfigure}{\textwidth}\centering
			\includegraphics[width=.9\textwidth]{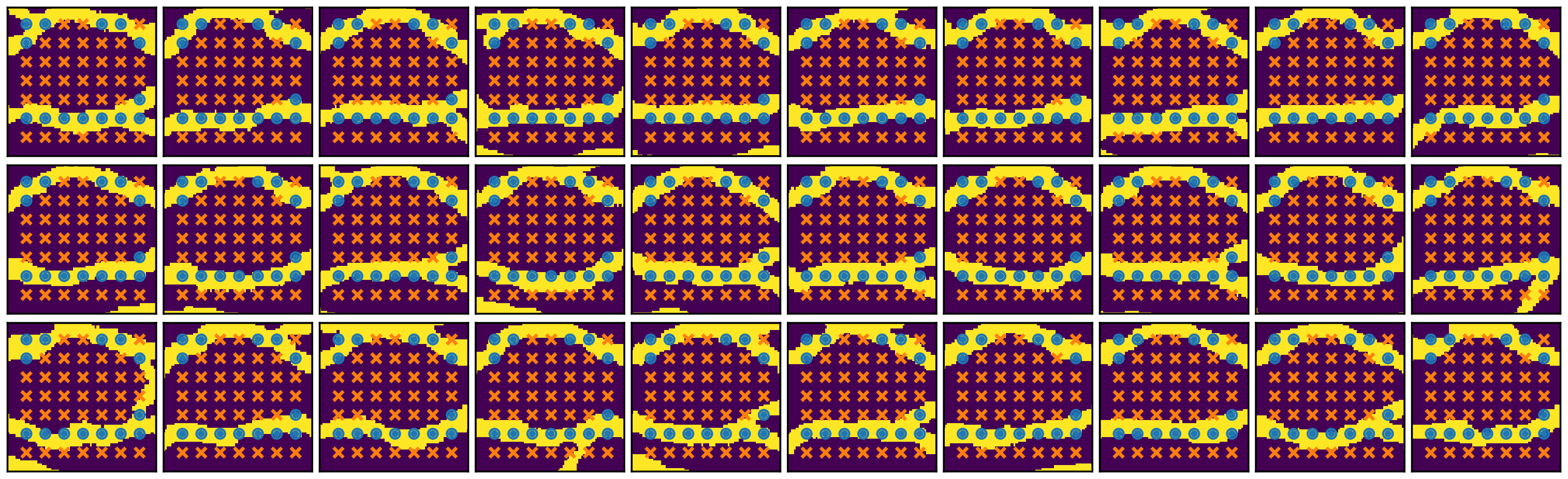}
			\caption{\texttt{snesim}\label{fig:snesim_cond21}}
		\end{subfigure}

		\begin{subfigure}{\textwidth}\centering
			\includegraphics[width=\textwidth]{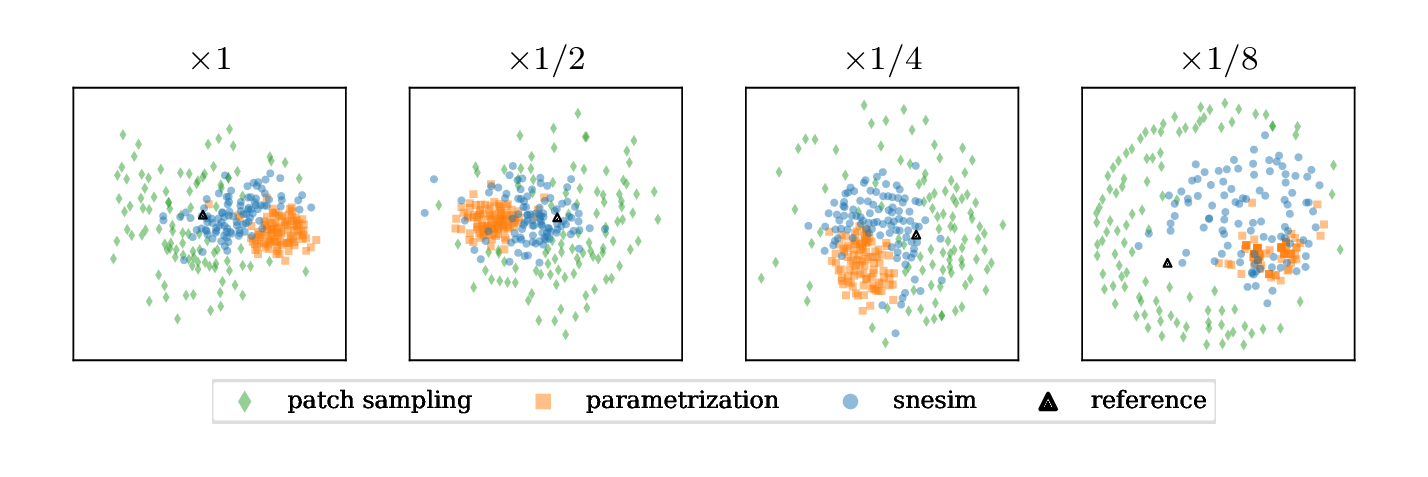}
			\vspace{-2em}\caption{Multidimensional scaling visualization.\label{fig:mds_cond21}}
		\end{subfigure}

		\caption{Example G. Conditional realizations.\label{fig:cond21}}
	\end{figure}

	\vspace{1em}
	\begin{table}[H]\centering
		\caption{Example G. ANODI scores (inconsistency/diversity).\label{table:cond21}}
		\includegraphics[width=\textwidth]{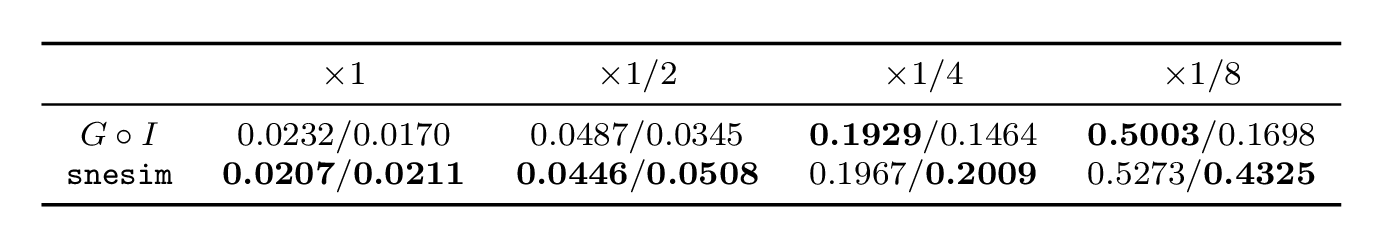}
		\vspace{-1.5em}
	\end{table}
\end{figure}

\begin{figure}\centering
	\begin{figure}[H]\centering
		\begin{subfigure}{\textwidth}\centering
			\includegraphics[width=.9\textwidth]{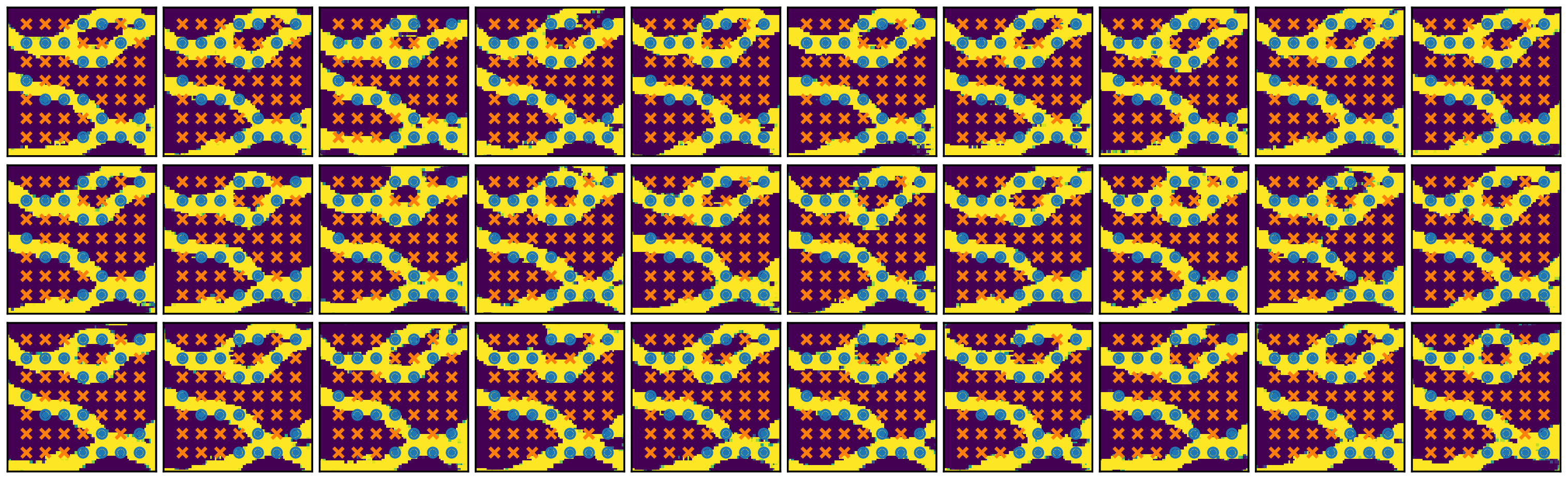}
			\caption{$G\circ I(\ww)$\label{fig:gen_cond22}}
		\end{subfigure}\vspace{1em}

		\begin{subfigure}{\textwidth}\centering
			\includegraphics[width=.9\textwidth]{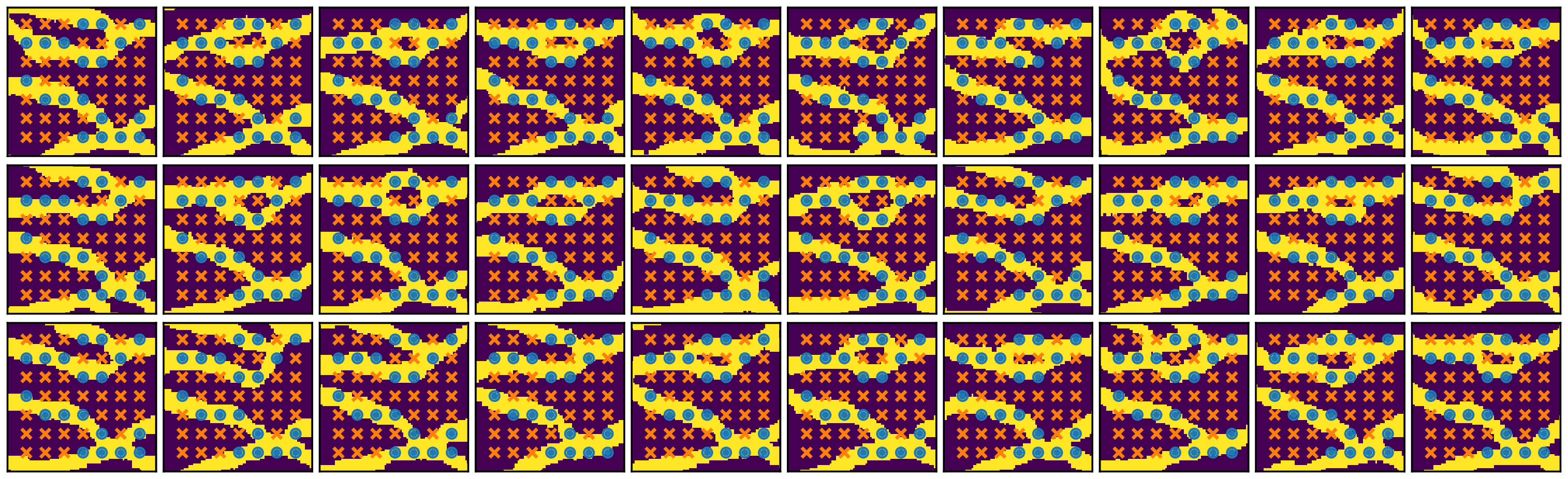}
			\caption{\texttt{snesim}\label{fig:snesim_cond22}}
		\end{subfigure}

		\begin{subfigure}{\textwidth}\centering
			\includegraphics[width=\textwidth]{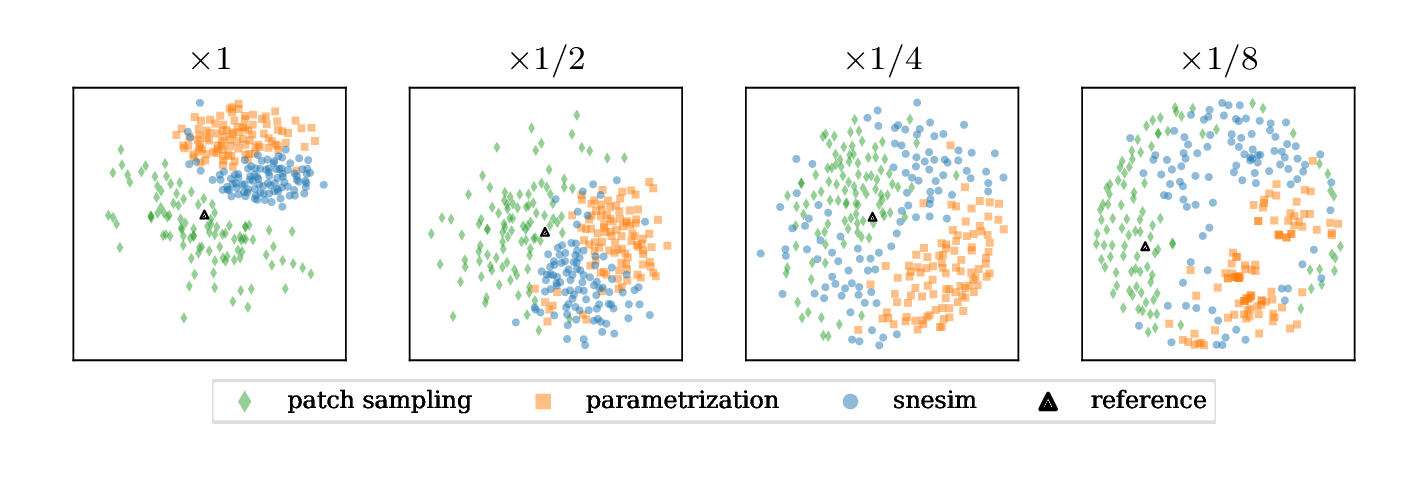}
			\vspace{-2em}\caption{Multidimensional scaling visualization and.\label{fig:mds_cond22}}
		\end{subfigure}

		\caption{Example H. Conditional realizations.\label{fig:cond22}}
	\end{figure}

	\vspace{1em}
	\begin{table}[H]\centering
		\caption{Example H. ANODI scores (inconsistency/diversity).\label{table:cond22}}
		\vspace{-1.5em}
		\includegraphics[width=\textwidth]{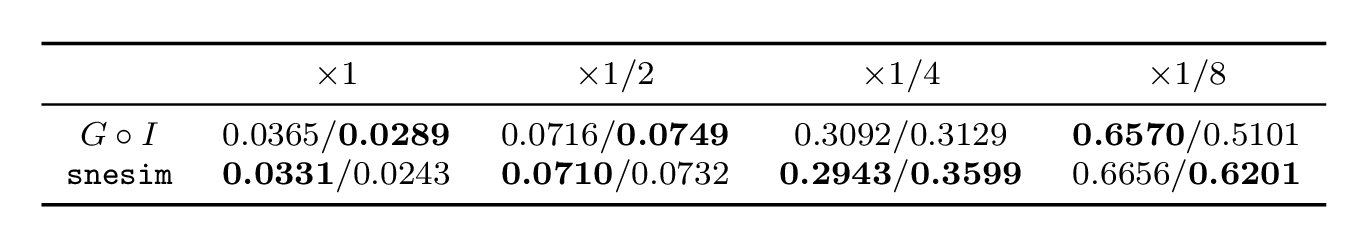}
	\end{table}
\end{figure}

\begin{figure}\centering
	\begin{figure}[H]\centering
		\begin{subfigure}{\textwidth}\centering
			\includegraphics[width=.9\textwidth]{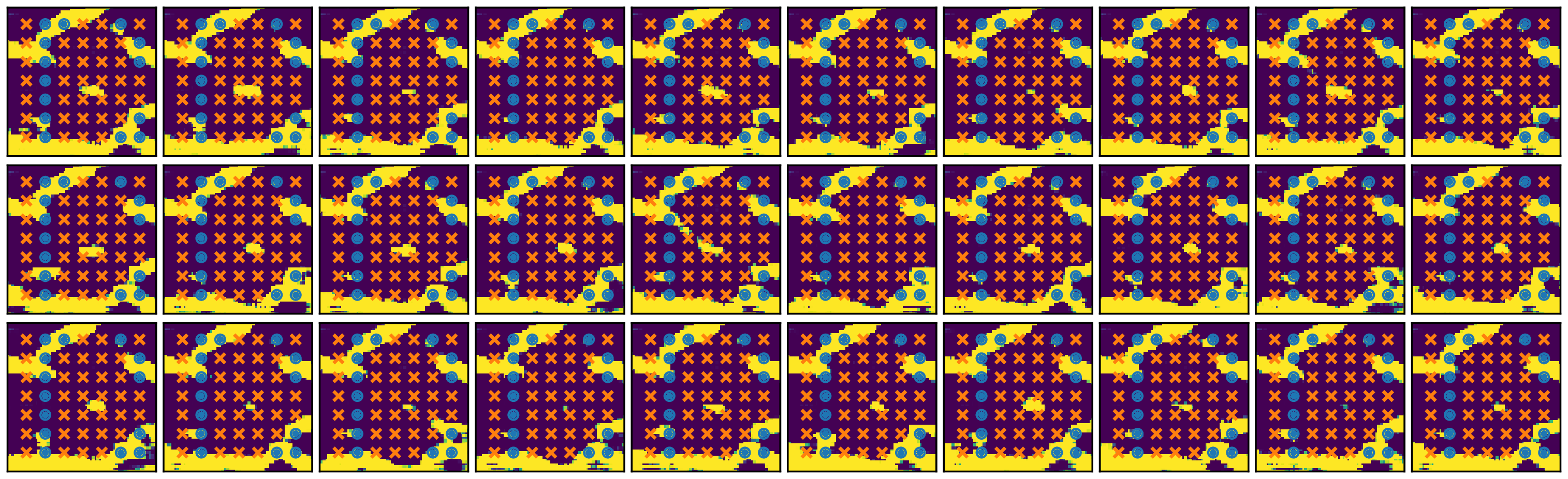}
			\caption{$G\circ I(\ww)$\label{fig:gen_cond23}}
		\end{subfigure}\vspace{1em}

		\begin{subfigure}{\textwidth}\centering
			\includegraphics[width=.9\textwidth]{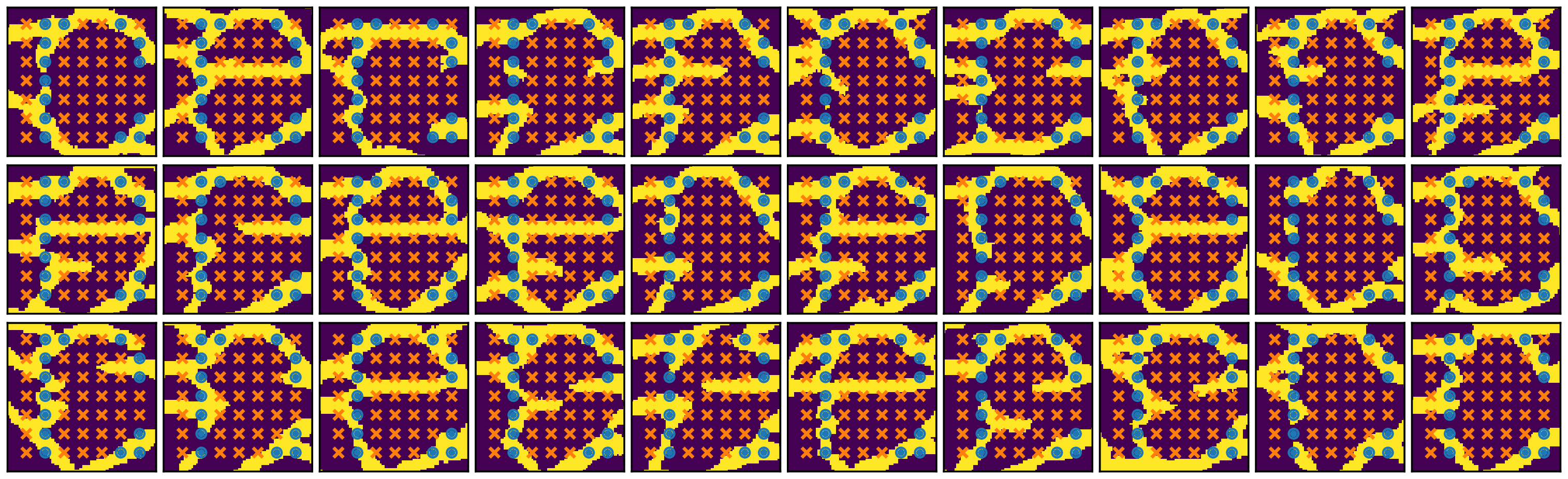}
			\caption{\texttt{snesim}\label{fig:snesim_cond23}}
		\end{subfigure}

		\begin{subfigure}{\textwidth}\centering
			\includegraphics[width=\textwidth]{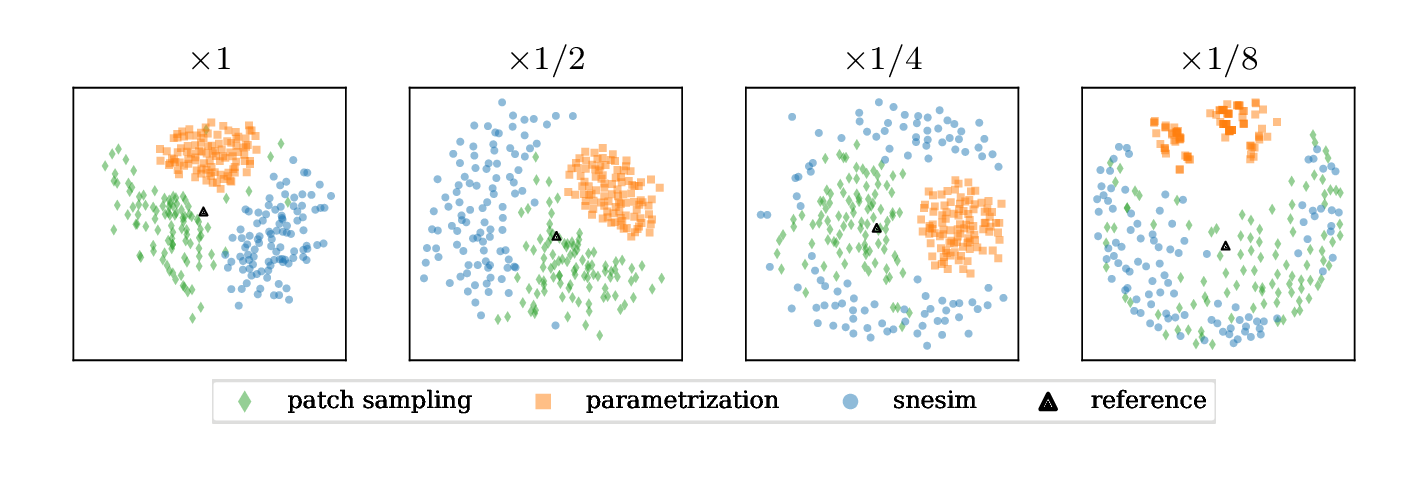}
			\vspace{-2em}\caption{Multidimensional scaling visualization.\label{fig:mds_cond23}}
		\end{subfigure}

		\caption{Example I. Conditional realizations.\label{fig:cond23}}
	\end{figure}

	\vspace{1em}
	\begin{table}[H]\centering
		\caption{Example I. ANODI scores (inconsistency/diversity).\label{table:cond23}}
		\vspace{-1.5em}
		\includegraphics[width=\textwidth]{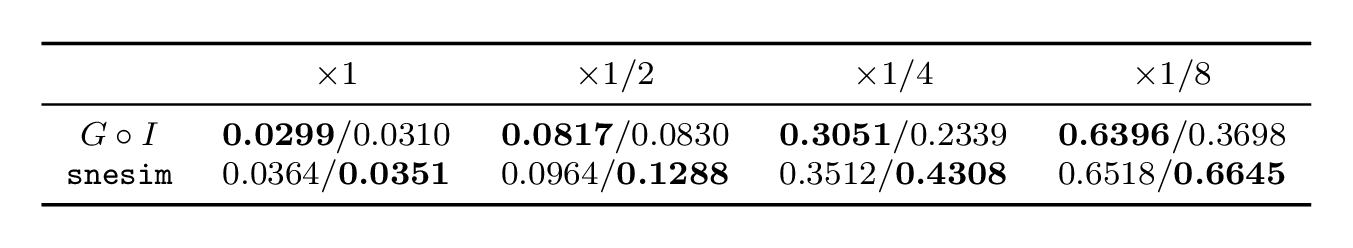}
	\end{table}
\end{figure}

\begin{figure}\centering
    \begin{subfigure}{.32\textwidth}\centering
        \includegraphics[width=\textwidth]{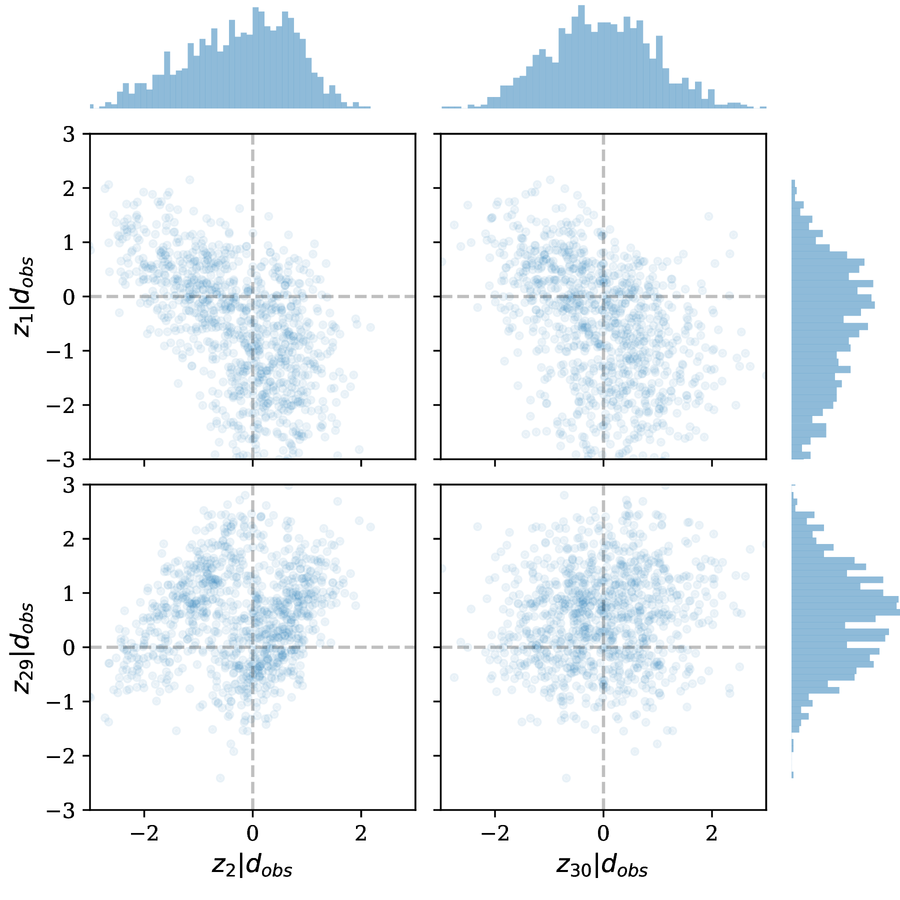}
        \caption{Example A}
    \end{subfigure}
    \begin{subfigure}{.32\textwidth}\centering
        \includegraphics[width=\textwidth]{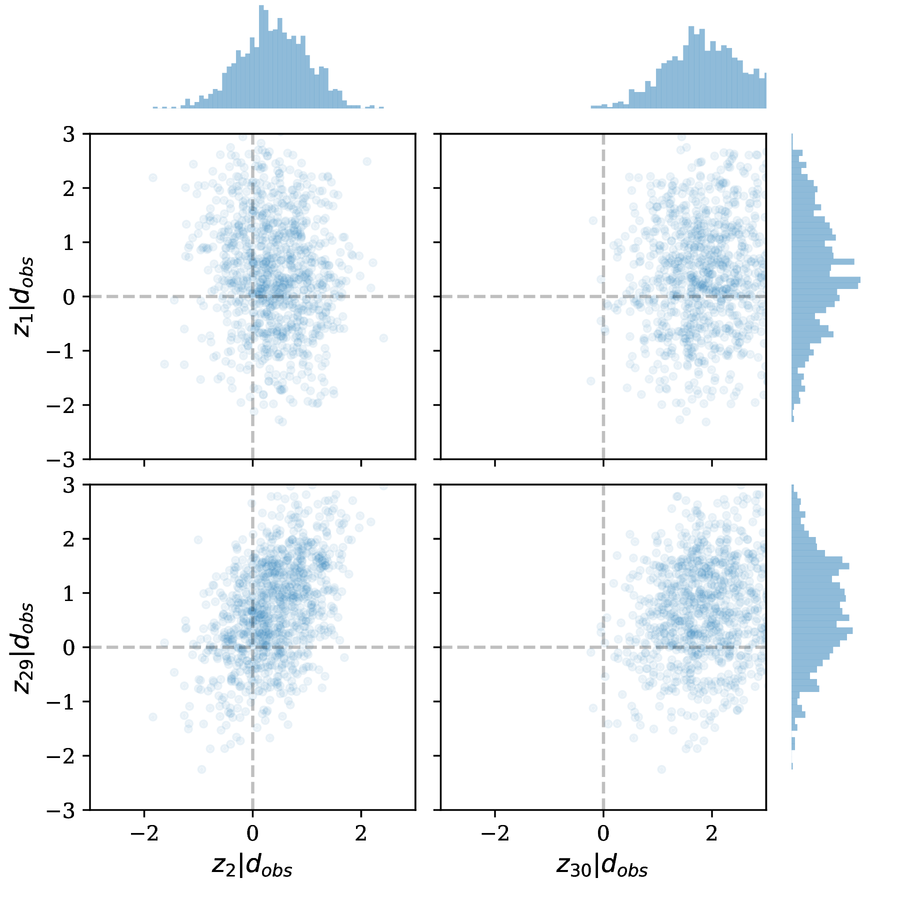}
        \caption{Example B}
    \end{subfigure}
    \begin{subfigure}{.32\textwidth}\centering
        \includegraphics[width=\textwidth]{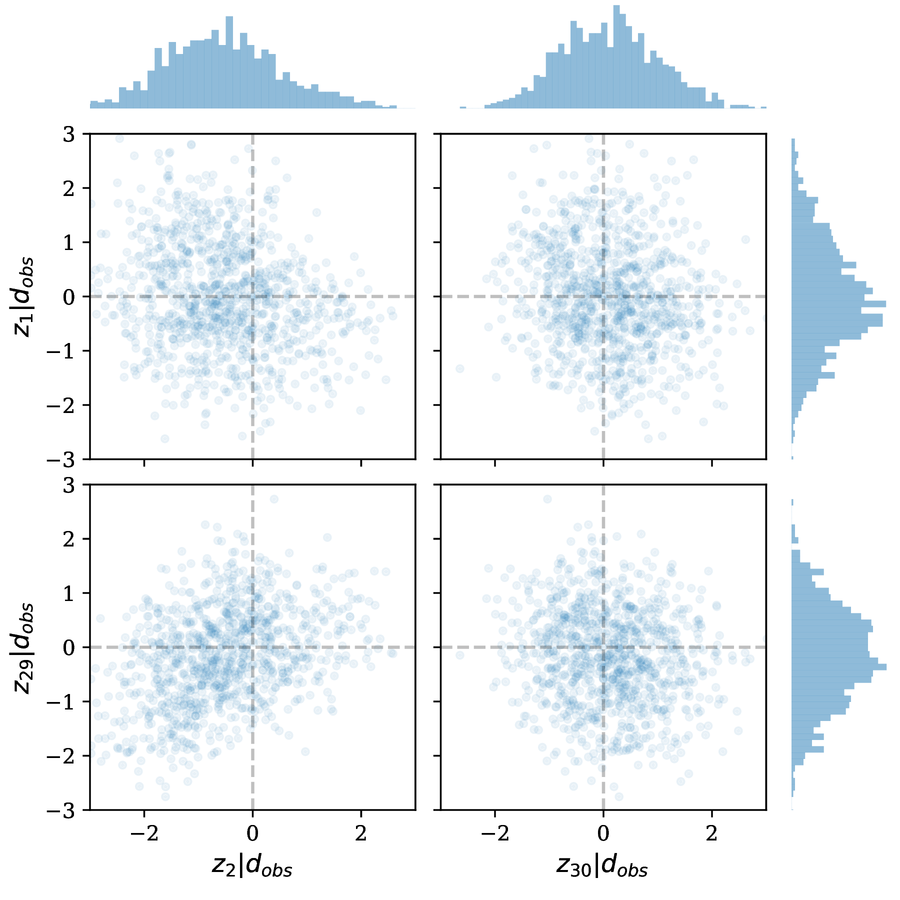}
        \caption{Example C}
    \end{subfigure}

    \vspace{1em}

    \begin{subfigure}{.32\textwidth}\centering
        \includegraphics[width=\textwidth]{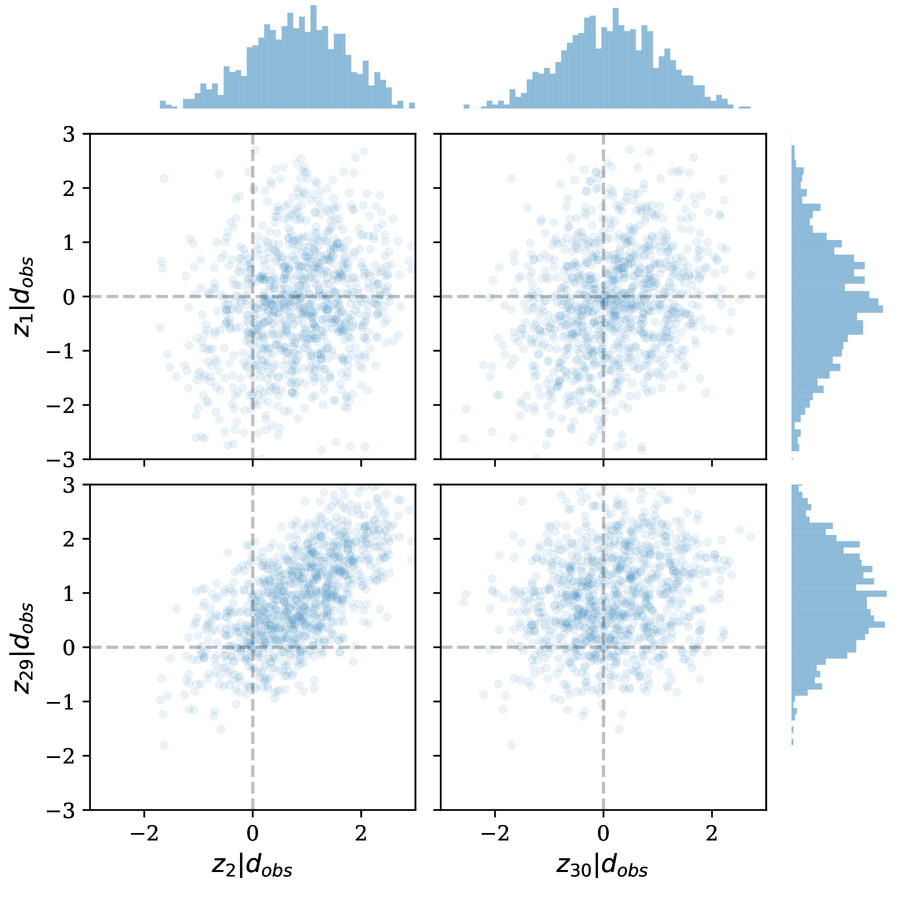}
        \caption{Example D}
    \end{subfigure}
    \begin{subfigure}{.32\textwidth}\centering
        \includegraphics[width=\textwidth]{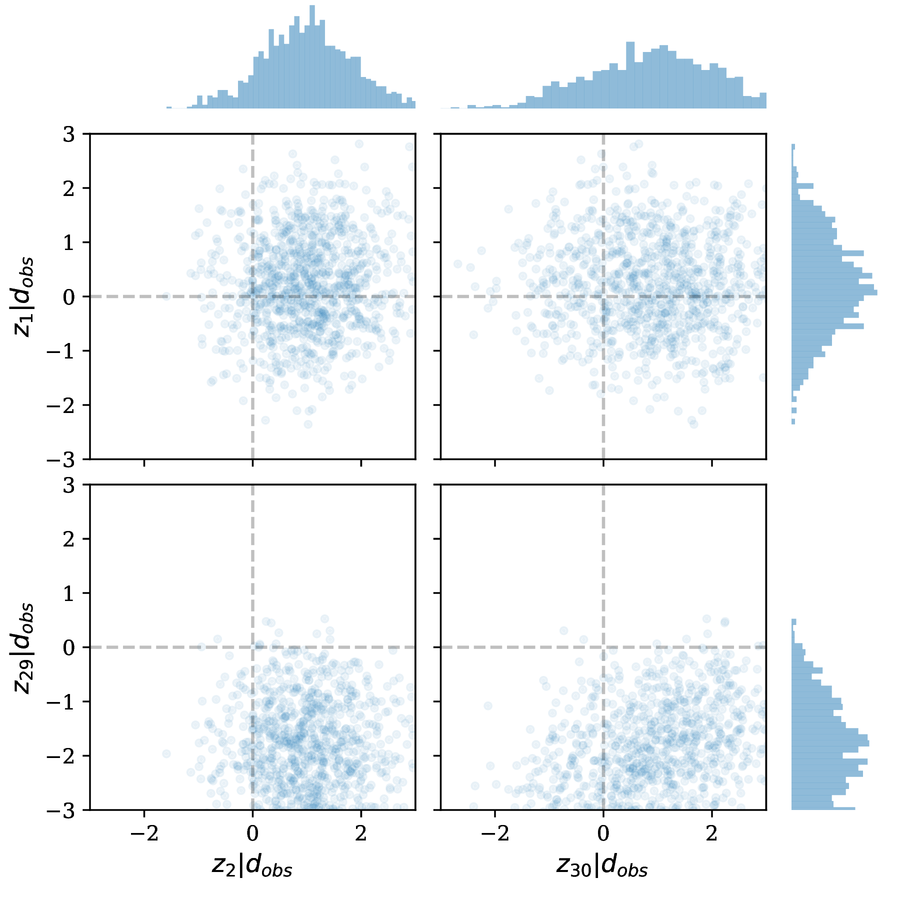}
        \caption{Example E}
    \end{subfigure}
    \begin{subfigure}{.32\textwidth}\centering
        \includegraphics[width=\textwidth]{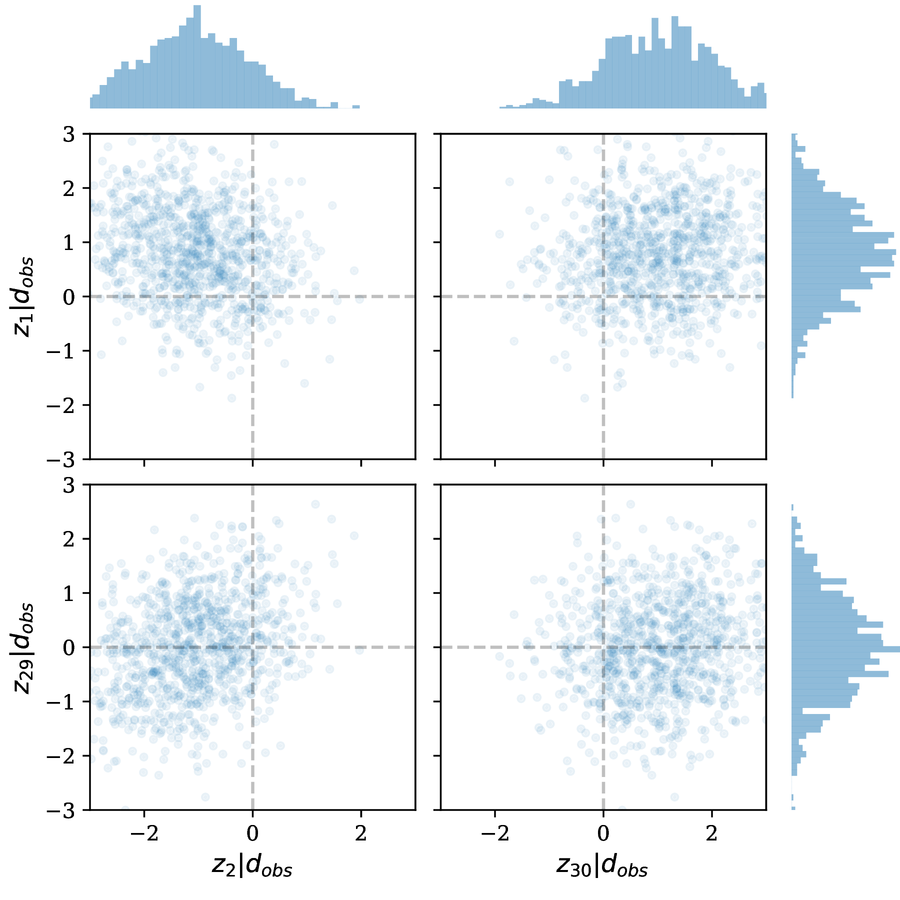}
        \caption{Example F}
    \end{subfigure}

    \vspace{1em}

    \begin{subfigure}{.32\textwidth}\centering
        \includegraphics[width=\textwidth]{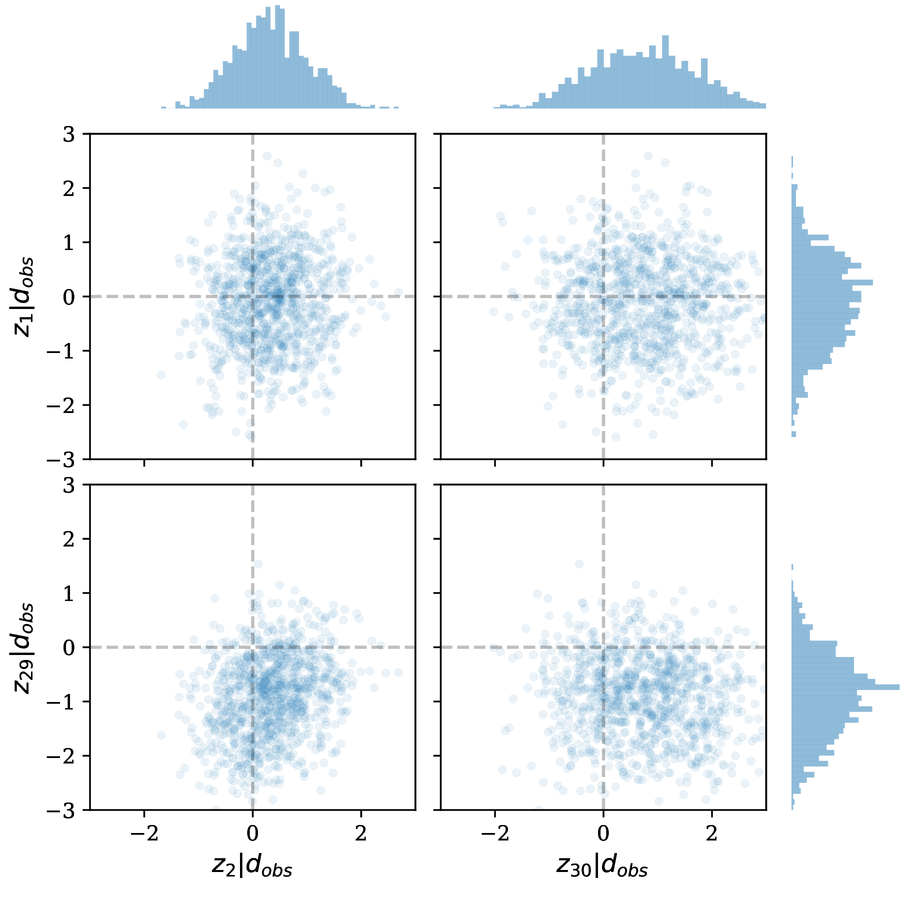}
        \caption{Example G}
    \end{subfigure}
    \begin{subfigure}{.32\textwidth}\centering
        \includegraphics[width=\textwidth]{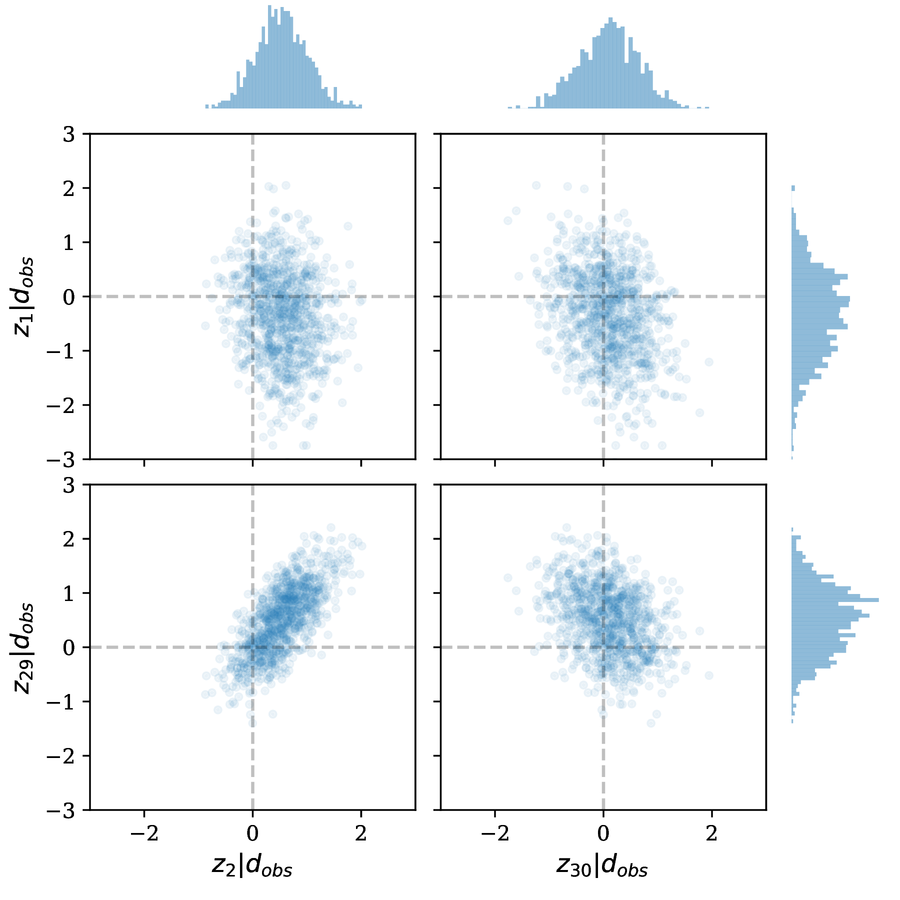}
        \caption{Example H}
    \end{subfigure}
    \begin{subfigure}{.32\textwidth}\centering
        \includegraphics[width=\textwidth]{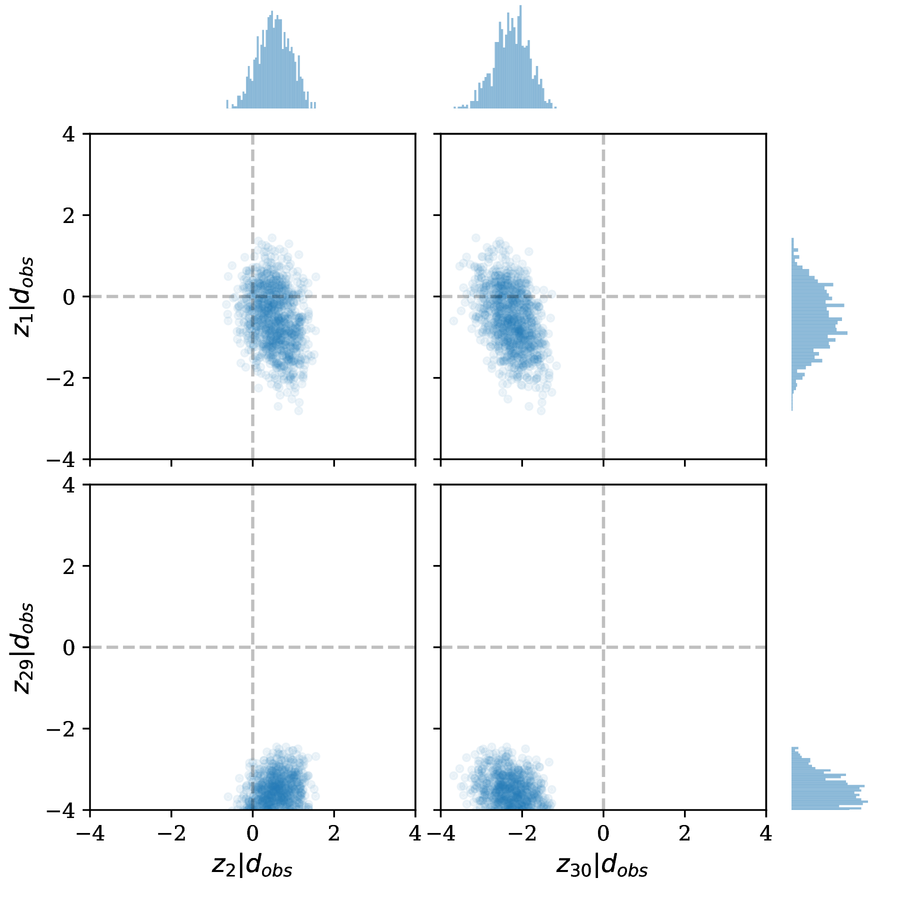}
        \caption{Example I}
    \end{subfigure}

    \caption{Visualizing the distribution of $\zz|d_{\mathrm{obs}}$.\label{fig:netI}}

\end{figure}

\begin{figure}
    \includegraphics[width=\textwidth]{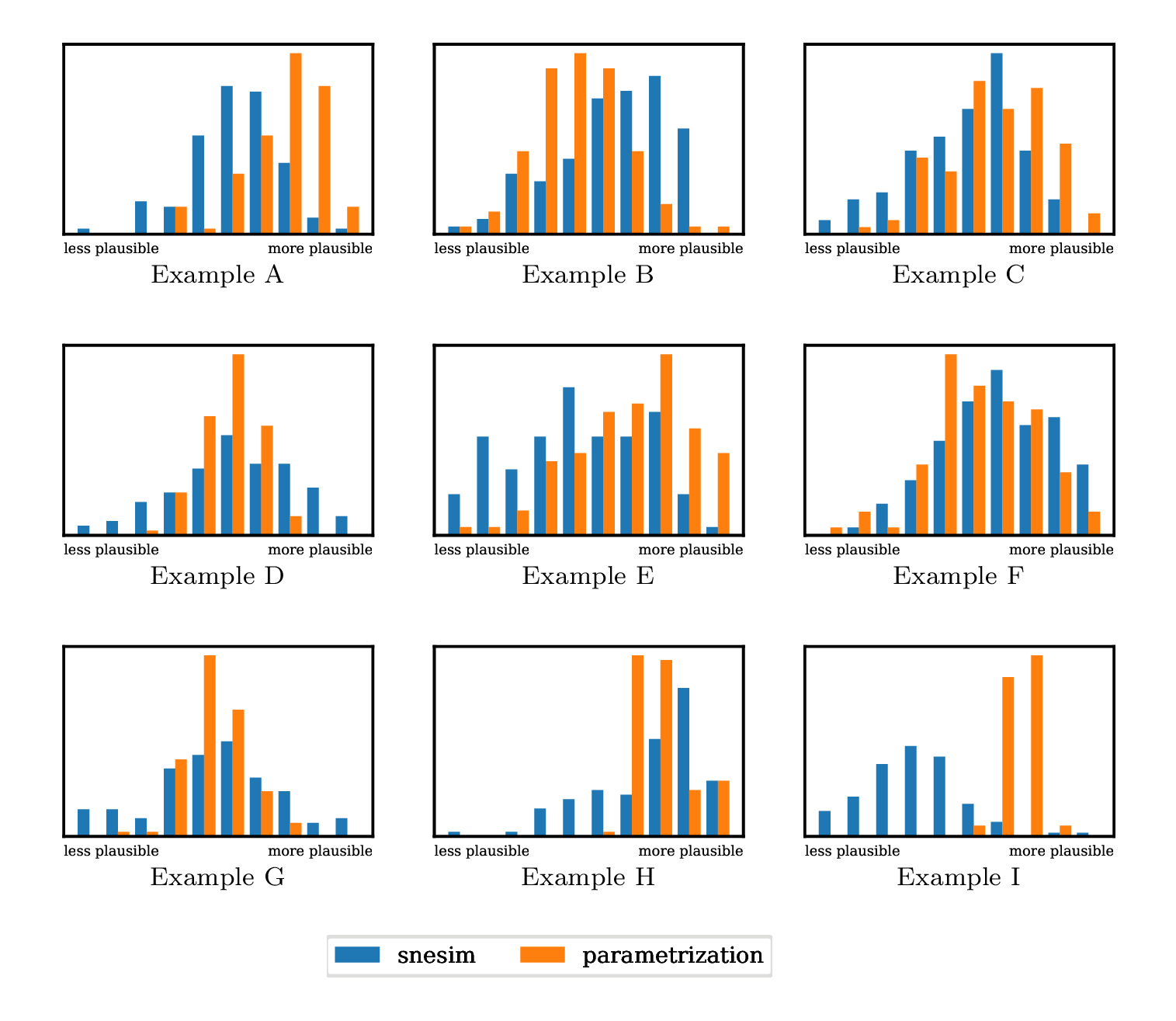}
    \caption{Histograms of discriminator scores.\label{fig:Deval}}
\end{figure}

We now obtain conditional generators for 9 conditioning configurations, ranging from 16 to 49 spatial observations (hard data). The configurations are detailed in~\Cref{table:conditioning}.
These indicate the presence or absence of channels at different locations of the domain.
For each configuration $d_{\mathrm{obs}}$, we derive the Bayesian posterior $p(z | d_{\mathrm{obs}})$ as described in~\Cref{sec:conditioning}, and train an inference network to sample from it as described in~\Cref{sec:methodology}.
We assume $\lambda=0.1$ in~\Cref{eq:lossfn} in all our test cases.
Note that we use a Gaussian likelihood function in the Bayesian formulation, although one could also consider the binomial distribution for binary data.
Also note that due to the Bayesian framework adopted here -- i.e. that the observations are noisy -- we cannot fully guarantee that conditioning is strictly honored, but only that it is honored with high probability.

\subsubsection{Architecture design -- inference network}
The inference network $I_\phi\colon \RR^{30}\to\RR^{30}$ is a fully connected neural network with several layers.
We naturally use $n_w=n_z=30$ and $p_w = p_z = \manN(\mathbf{0},\mathbf{I})$ (so that if no conditioning were present, $I_\phi$ should learn a distribution-preserving function such as the identity function).
To simplify the presentation, we perform minimal hyperparameter tuning -- that is, we use the same neural network architecture and optimization parameters for all 9 test cases. In practice, one should perform hyperparameter optimization for each problem at hand.
For the non-linearity, we use scaled exponential linear units~\citep{klambauer2017self}. No non-linearity is applied in the output layer.
Further details of the architecture and training are given in~\Cref{sec:details_inf}.
Once $I$ is trained, we use $G\circ I$ to generate conditional realizations.

\subsubsection{Quality assessment -- conditional realizations}\label{sec:quality_cond}
We show conditional realizations generated by $G\circ I$ for each test case
in~\Cref{fig:cond01,fig:cond02,fig:cond03,fig:cond11,fig:cond12,fig:cond13,fig:cond21,fig:cond22,fig:cond23}.
We also include conditional realizations obtained using \texttt{snesim} for comparison.
Over the images, we indicate the conditioning using blue dots to denote channel material and orange crosses to denote background material.
Overall, we observe that $G\circ I$ generates good conditioning results maintaining the plausibility of the realizations.
We also show in~\Cref{fig:netI} the output of the inference network $I$ for each test case to visualize the distribution change (for no conditioning, the distribution is normal). Since it is cumbersome to visualize the distribution for the 30 components of $\zz$, we show pairwise scatter plots only for the first and last two components.

\paragraph{Assessment using analysis of distances}
We perform a quantitative assessment as in the unconditional case, using the ANODI method and multidimensional scaling on sets of 100 realizations.
We keep the ``patch sampling'' method for the multidimensional scaling visualization.
The results are shown in~\Cref{fig:cond01,fig:cond02,fig:cond03,fig:cond11,fig:cond12,fig:cond13,fig:cond21,fig:cond22,fig:cond23} and~\Cref{table:cond01,table:cond02,table:cond03,table:cond11,table:cond12,table:cond13,table:cond21,table:cond22,table:cond23}.
In terms of the ANODI scores, we find that whenever one method generates more plausible images (lower inconsistency), it also tends to be less diverse, and vice versa -- this is the usual trade-off in image synthesis.
Overall, we find that \texttt{snesim} produces more diverse realizations whereas $G\circ I$ emphasizes on plausibility.
This is reasonable since $G\circ I \subset G$, i.e. an output of $G\circ I$ is an output of $G$, therefore the conditional realizations cannot deviate too much from the reference spatial statistics.
Also for this reason, we find that the outputs of $G\circ I$ and \texttt{snesim} are most different when the conditioning statistics are in less agreement with the reference spatial statistics.
This is evident in~\Cref{fig:cond23}, and to a lesser extent~\Cref{fig:cond01,fig:cond03}.
In~\Cref{fig:cond03} we enforce a diagonal channel, finding that $G\circ I$ generates plausible but noticeably less diverse realizations compared to \texttt{snesim}.
The difference is more pronounced in~\Cref{fig:cond23} where we densely enforce vertical channels and find a failure case for $G\circ I$, whereas \texttt{snesim} can handle this case despite the implausibility of this conditioning (there are no vertical channels in the reference image).
In other words, if the conditioning is in far disagreement with the reference spatial statistics, effective conditional parametrization may be difficult since $G$ is tied to the reference statistics.
In the \texttt{snesim} algorithm, deliberate conditioning and diversity can be achieved regardlessly since the conditioning is trivially imposed and the stochasticity is intrinsic to the synthesis process.

Finally, when the conditioning is in good agreement with the reference spatial statistics as in the remaining cases, we observe that $G\circ I$ generates realizations that are visually comparable with \texttt{snesim}.
Note that in practice, we always aim to use a reference image whose spatial statistics are in good agreement with the spatial observations, otherwise the reference image may not be representative of the area under study.
Also note that although we compare our method against a multipoint geostatistical simulator, our emphasis is on parametrization.
Lastly, we mention that the present results could be further improved with hyperparameter tuning for each individual test case.

\paragraph{Assessment using the discriminator}
We demonstrate an alternative approach to assess the quality of the generated realizations using the discriminator $D$ that is made available after training generative adversarial networks to obtain $G$. Recall that the discriminator outputs a score that estimates the probability of a realization being ``real'' (see~\Cref{sec:gan}), with higher scores corresponding to higher probability. We can therefore use the discriminator to assess the quality of the generated realizations.
We evaluate the discriminator on the same sets of 100 realizations used before in the ANODI assessment, and plot the histogram of the scores in~\Cref{fig:Deval}.
Overall, we verify that this assessment at least arrives at the same qualitative conclusions as the ANODI assessment. For a quantitative summary report, one can consider summary statistics of the scores such as the mean and variance as measures of plausibility and diversity, respectively.
Another option is to report the Jensen-Shannon divergence with respect to some reference histogram.

\section{Related work} \label{sec:related}

There is increasing interest in applying deep learning techniques in geological applications to leverage recent advances in the field as well as the increasing availability of data and computational resources that make these techniques effective.
In particular, we expect to see more applications of generative adversarial networks (GAN)~\citep{goodfellow2014generative} in geology following successful results from recent works~\citep{mosser2017reconstruction,mosser2017stochastic,chan2017parametrization,laloy2018training,dupont2018generating,mosser2018conditioning}.
In~\citep{laloy2018training}, conditioning is addressed using a Bayesian formulation and performing Markov chain Monte Carlo to sample the corresponding posterior distribution. In~\citep{dupont2018generating,mosser2018conditioning}, the authors address conditioning using the inpainting technique from~\citep{yeh2016semantic}, which is equivalent to a Bayesian formulation using a ``neural network prior'' (see~\Cref{sec:inpainting} for more details), and the sampling is done using local optimization.
Our approach is closer to~\citep{laloy2018training} in that we use a simple prior and aim to sample the full posterior, with the difference that the sampling is carried out by a neural network and we obtain a parametrization for the sampling process.
Our approach is motivated by~\citep{ulyanov2017improved} where the authors trained a neural network to perform texture synthesis.
Such authors used the sample entropy estimator for case $k=1$ (nearest neighbor, see~\Cref{eq:kozachenko}).
The entropy estimator used in our work is a generalization introduced in~\citep{goria2005new}.
A similar estimator based on random distances is used in~\citep{li2017diversified} in the context of texture synthesis.
In the context of generative modeling, \citep{kim2016deep} used a closed-form expression of the entropy term when using batch normalization~\citep{ioffe2015batch}.
Other alternatives to train neural samplers include normalizing flow~\citep{rezende2015variational}, autoregressive flow~\citep{kingma2016improved}, and Stein discrepancy~\citep{wang2016learning}.
These are all alternatives worth exploring in future work.
Also related to our work include~\citep{nguyen2016plug,engel2017latent} where the authors optimize the latent space to condition on labels/classes.

\section{Conclusion} \label{sec:conclusion}

We introduced a method to obtain a conditional parametrization by extending an existing unconditional parametrization, enabling reusability as well as  direct and parametric sampling of conditional realizations.
The parametrization considered in this work was based on deep neural networks motivated by their ability to express complex high-dimensional data such as natural images, including geological subsurface images.
The unconditional parametrization $G$ was obtained using generative adversarial networks (GAN)~\citep{goodfellow2014generative}, and the post-hoc conditioning was done by training a second neural network $I$ to sample a Bayesian posterior, resulting in $G\circ I$ as the conditional parametrization.

We applied the method to parametrize binary channelized images using the benchmark image of~\citet{strebelle2001reservoir}.
In previous works, unconditional parametrization based on GAN was assessed using mostly two-point statistics tools.
Here we added to the assessment using the analysis of distances method~\citep{tan2014comparing} which captures multipoint statistics.
We found very positive results for the unconditional case, supporting previous results showing that $G$ can effectively replicate the data generating process (in our case, the \texttt{snesim}~\citep{strebelle2001reservoir} algorithm) while achieving dimensionality reduction of two orders of magnitude.
Post-hoc conditional parametrization was explored for a variety of configurations. We found that $G\circ I$ produces very plausible realizations with good conditioning results, but the effectiveness may depend on the conditioning.
Specifically, if the observations are in far disagreement with the reference spatial statistics, effective conditioning may be difficult.
For observations that agree with the reference spatial statistics, we found that the parametrization produces comparable results.
% Our results could be improved with further hyperparameter tuning for each individual test case.
%

Possible future works include studying  alternative training methods for the inference network as mentioned in~\Cref{sec:related}, and further assessments with other images and in large scale settings.
% Possible future works include using a binomial distribution for the Bayesian likelihood, alternative training methods for the inference network as mentioned in~\Cref{sec:related}, and assessments with other images.

\bibliographystyle{unsrtnat}
\bibliography{biblio}

\appendix

\section{Implementation details}\label{sec:details}

This section describes training and hyperparameters of the neural network models. See~\citep{bengio2012practical} for a practical guide on training neural networks.

\begin{table}\centering\small
	\begin{subtable}[t]{.45\textwidth}\centering
    \begin{tabular}{l l}
        \toprule
        State size               & Layer                            \\
        \midrule
        $30 \times  1\times 1 $  & \texttt{ConvT(4,1,0), BN, ReLU}  \\
        $512\times  4\times 4 $  & \texttt{ConvT(4,2,1), BN, ReLU} \\
        $256\times  8\times 8 $  & \texttt{ConvT(4,2,1), BN, ReLU} \\
        $128\times 16\times 16$  & \texttt{ConvT(4,2,1), BN, ReLU} \\
        $64 \times 32\times 32$  & \texttt{ConvT(4,2,1), Tanh} \\
        $1  \times 64\times 64$  & -- \\
        \bottomrule
    \end{tabular}
    \caption{Generator architecture\label{table:generator}.}
    \end{subtable}\hfill
    \begin{subtable}[t]{.45\textwidth}\centering
    \begin{tabular}{l l}
        \toprule
        State size  & Layer \\
        \midrule
        $30  $  & \texttt{FC, SeLU}  \\
        $512 $  & \texttt{FC, SeLU} \\
        $\vdots$ & $\vdots$ \\
        % $512 $  & \texttt{FC, SeLU} \\
        % $512 $  & \texttt{FC, SeLU} \\
        % $512 $  & \texttt{FC, SeLU} \\
        % $512 $  & \texttt{FC, SeLU} \\
        $512 $  & \texttt{FC, SeLU} \\
        $512 $  & \texttt{FC} \\
        $30  $  & -- \\
        \bottomrule
    \end{tabular}
    \caption{Inference network architecture\label{table:inference}}
    \end{subtable}
	\caption{Neural network parametrization.
        \texttt{ConvT}=transposed convolution, the triplet indicates (filter size, stride, padding). \texttt{BN}=batch normalization. \texttt{FC}=fully connected.}
\end{table}

\subsection{Generator neural network}\label{sec:details_gen}

The generator $G\colon\RR^{30}\to\RR^{64\times64}$ is a deep convolutional neural network based on the template provided in~\citep{radford2015unsupervised}.
The generator architecture consists of stacks of (transposed) convolutional layers (see~\Cref{sec:cnn})
together with batch normalization layers~\citep{ioffe2015batch}.
Batch normalization is the operation of normalizing the intermediate layer results to have zero mean and unit variance, which drastically improves optimization of deep neural networks~\citep{ioffe2015batch}.
For the non-linearity, we use rectified linear units (ReLU, $\sigma(x)=\max(0,x)$) in the intermediate layers, and $\sigma(x)=\tanh(x)$ in the last layer to constrain the output in $[-1,1]$.
The architecture is summarized in~\Cref{table:generator}.
We train $G$ using the Wasserstein formulation of GAN introduced in~\citep{arjovsky2017wasserstein}
with the proposed default hyperparameters. The optimization is performed using the
Adam~\citep{kingma2014adam,reddi2018convergence} method with
learning rate of $10^{-4}$ and batch size of $32$.
Our generator converges in approximately \num{20000} iterations, taking around $30$ minutes using a Nvidia GeForce GTX Titan X GPU.
For deployment, it can generate approximately \num{5500} realizations per second using the GPU.

\subsection{Inference neural network}\label{sec:details_inf}

We use the same inference network architecture $I\colon\RR^{30}\to\RR^{30}$ for all our conditioning experiments.
The architecture is simply a stack of fully connected layers with constant-size intermediate layers.
More specifically, we first transform the input from size 30 to size 512, then apply several more intermediate transformations preserving the size, and finally apply a transformation to bring the size back from 512 to 30 in the output layer.
For the non-linearity, we use scaled exponential linear units (SeLU)~\citep{klambauer2017self}, which are the current default option for deep fully connected networks: $\sigma(x) = \lambda x$ if $x>0$, otherwise $\sigma(x) = \lambda \alpha (e^x - 1)$, where constants $\lambda, \alpha$ are given in~\citep{klambauer2017self}.
No non-linearity is applied in the output layer (we do not need to bound the output as in the case of the generator).
We experimented with different numbers of layers. Perhaps not surprisingly, we found that deeper architectures tended to produce better results in general. In our work, we settled with 5 intermediate layers. The architecture is summarized in~\Cref{table:inference}.
We optimize $I$ using the Adam method with learning rate of $10^{-4}$ and batch size of 64 for all the test cases.
The network converges in between \num{1000} and \num{10000} iterations, depending on the conditioning, taking between seconds and a few minutes to train using a Nvidia GeForce GTX Titan X GPU.
For deployment, the conditional generator $G\circ I$ can generate approximately \num{5500} realizations per second using the GPU -- we do not see significant increase in generation time from $G$ to $G\circ I$.

\section{Conditioning settings}
The conditioning settings are summarized in~\Cref{table:conditioning}.

% \rowcolors{3}{gray}{white}
\begin{table}\centering \tiny

    \begin{subtable}{\textwidth}\centering
    \begin{tabular}{*{18}c}
        \toprule
        \multicolumn{3}{c}{A} & \multicolumn{3}{c}{B} & \multicolumn{3}{c}{C} & \multicolumn{3}{c}{D} & \multicolumn{3}{c}{E} & \multicolumn{3}{c}{F} \\

        $i$ & $j$ & val & $i$ & $j$ & val & $i$ & $j$ & val & $i$ & $j$ & val & $i$ & $j$ & val & $i$ & $j$ & val \\
        \midrule

        12 & 12 & 0 & 12 & 12 & 1 & 12 & 12 & 1 &  0 & 50 & 1 &  0 & 20 & 1 & 33 & 44 & 1 \\
        12 & 25 & 0 & 25 & 12 & 0 & 25 & 12 & 0 & 10 & 50 & 1 &  5 & 22 & 1 & 28 & 42 & 1 \\
        12 & 38 & 1 & 38 & 12 & 0 & 38 & 12 & 0 & 20 & 50 & 1 & 10 & 23 & 1 & 24 & 40 & 1 \\
        12 & 51 & 1 & 51 & 12 & 1 & 51 & 12 & 0 & 30 & 50 & 1 & 15 & 25 & 1 & 18 & 35 & 1 \\
        25 & 12 & 1 & 12 & 25 & 0 & 12 & 25 & 0 & 40 & 50 & 1 & 20 & 26 & 1 & 16 & 30 & 1 \\
        25 & 25 & 0 & 25 & 25 & 1 & 25 & 25 & 1 & 50 & 50 & 1 & 30 & 30 & 1 & 20 & 23 & 1 \\
        25 & 38 & 0 & 38 & 25 & 1 & 38 & 25 & 0 & 60 & 50 & 1 & 40 & 33 & 1 & 27 & 20 & 1 \\
        25 & 51 & 0 & 51 & 25 & 0 & 51 & 25 & 0 &  0 & 15 & 1 & 45 & 34 & 1 & 32 & 19 & 1 \\
        38 & 12 & 0 & 12 & 38 & 0 & 12 & 38 & 0 & 10 & 21 & 1 & 50 & 36 & 1 & 39 & 21 & 1 \\
        38 & 25 & 1 & 25 & 38 & 1 & 25 & 38 & 0 & 20 & 26 & 1 & 55 & 37 & 1 & 45 & 24 & 1 \\
        38 & 38 & 1 & 38 & 38 & 1 & 38 & 38 & 1 & 30 & 32 & 1 & 60 & 39 & 1 & 48 & 32 & 1 \\
        38 & 51 & 1 & 51 & 38 & 0 & 51 & 38 & 0 & 40 & 37 & 1 & 60 & 20 & 1 & 43 & 37 & 1 \\
        51 & 12 & 0 & 12 & 51 & 1 & 12 & 51 & 0 & 50 & 43 & 1 & 55 & 22 & 1 & 36 & 40 & 1 \\
        51 & 25 & 0 & 25 & 51 & 0 & 25 & 51 & 0 & 60 & 48 & 1 & 50 & 23 & 1 &    &    &   \\
        51 & 38 & 0 & 38 & 51 & 0 & 38 & 51 & 0 & 10 & 15 & 1 & 45 & 25 & 1 &    &    &   \\
        51 & 51 & 1 & 51 & 51 & 1 & 51 & 51 & 1 & 20 & 15 & 1 & 40 & 26 & 1 &    &    &   \\
           &    &   &    &    &   &    &    &   & 30 & 15 & 1 & 35 & 30 & 1 &    &    &   \\
           &    &   &    &    &   &    &    &   & 40 & 15 & 1 & 20 & 33 & 1 &    &    &   \\
           &    &   &    &    &   &    &    &   & 50 & 15 & 1 & 15 & 34 & 1 &    &    &   \\
           &    &   &    &    &   &    &    &   & 60 & 15 & 1 & 10 & 36 & 1 &    &    &   \\
           &    &   &    &    &   &    &    &   &    &    &   &  5 & 37 & 1 &    &    &   \\
           &    &   &    &    &   &    &    &   &    &    &   &  0 & 39 & 1 &    &    &   \\

        \bottomrule
    \end{tabular}
    \caption{A-F}
    \end{subtable}\vspace{1em}

% \end{table}

% \begin{table}\centering \tiny
    \begin{subtable}[t]{\textwidth}\centering
    \begin{tabular}{r c c c c c c c}
        \toprule
            & $j=8$ & $j=16$ & $j=24$ & $j=32$ & $j=40$ & $j=48$ & $j=56$ \\
        \midrule
        $i= 8$ & 0 & 0 & 0 & 0 & 0 & 0 & 0 \\
        $i=16$ & 1 & 1 & 1 & 1 & 1 & 1 & 1 \\
        $i=24$ & 0 & 0 & 0 & 0 & 0 & 0 & 1 \\
        $i=32$ & 0 & 0 & 0 & 0 & 0 & 0 & 0 \\
        $i=40$ & 0 & 0 & 0 & 0 & 0 & 0 & 0 \\
        $i=48$ & 1 & 0 & 0 & 0 & 0 & 0 & 1 \\
        $i=56$ & 1 & 1 & 0 & 0 & 1 & 1 & 0 \\
        \bottomrule
    \end{tabular}
    \caption{G}
    \end{subtable}\vspace{1em}

    \begin{subtable}[t]{\textwidth}\centering
    \begin{tabular}{r c c c c c c c}
        \toprule
            & $j=8$ & $j=16$ & $j=24$ & $j=32$ & $j=40$ & $j=48$ & $j=56$ \\
        \midrule
        $i= 8$ & 0 & 0 & 0 & 1 & 1 & 1 & 1 \\
        $i=16$ & 0 & 0 & 0 & 0 & 1 & 0 & 1 \\
        $i=24$ & 0 & 1 & 1 & 1 & 0 & 0 & 0 \\
        $i=32$ & 1 & 0 & 0 & 0 & 0 & 0 & 0 \\
        $i=40$ & 0 & 0 & 0 & 1 & 1 & 0 & 0 \\
        $i=48$ & 1 & 1 & 1 & 0 & 0 & 1 & 0 \\
        $i=56$ & 0 & 0 & 0 & 1 & 1 & 0 & 1 \\
        \bottomrule
    \end{tabular}
    \caption{H}
    \end{subtable}\vspace{1em}

    \begin{subtable}[t]{\textwidth}\centering
    \begin{tabular}{r c c c c c c c}
        \toprule
            & $j=8$ & $j=16$ & $j=24$ & $j=32$ & $j=40$ & $j=48$ & $j=56$ \\
        \midrule
        $i= 8$ & 0 & 1 & 0 & 0 & 0 & 1 & 1 \\
        $i=16$ & 0 & 1 & 0 & 0 & 0 & 0 & 1 \\
        $i=24$ & 0 & 1 & 0 & 0 & 0 & 0 & 0 \\
        $i=32$ & 0 & 1 & 0 & 0 & 0 & 0 & 0 \\
        $i=40$ & 0 & 1 & 0 & 0 & 0 & 0 & 1 \\
        $i=48$ & 0 & 1 & 0 & 0 & 0 & 0 & 1 \\
        $i=56$ & 0 & 1 & 1 & 0 & 0 & 1 & 0 \\
        \bottomrule
    \end{tabular}
    \caption{I}
    \end{subtable}

    \caption{Conditioning configuration for each test case. The pair $(i,j)$ denotes cell indices (row and column, respectively), and $\mathrm{val}=1$ indicates channel material, while $\mathrm{val}=0$ indicates background material.\label{table:conditioning}}

\end{table}

\section{Mixture of Gaussians} \label{sec:mixture}

\begin{figure}\centering
  \begin{subfigure}{.8\textwidth}\centering
  \includegraphics[width=.49\textwidth]{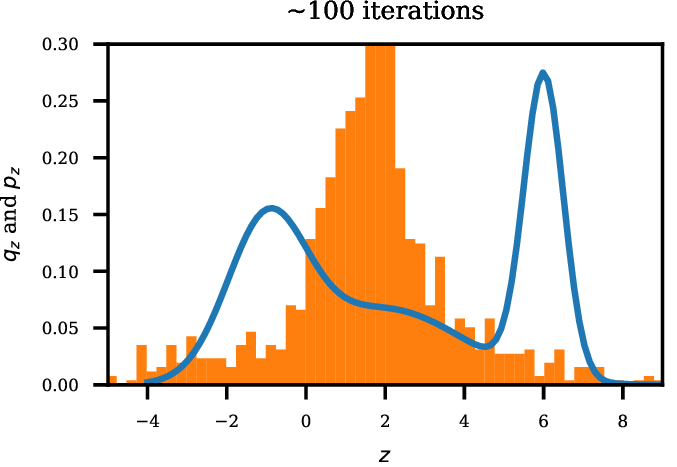}
  \hfill
  \includegraphics[width=.49\textwidth]{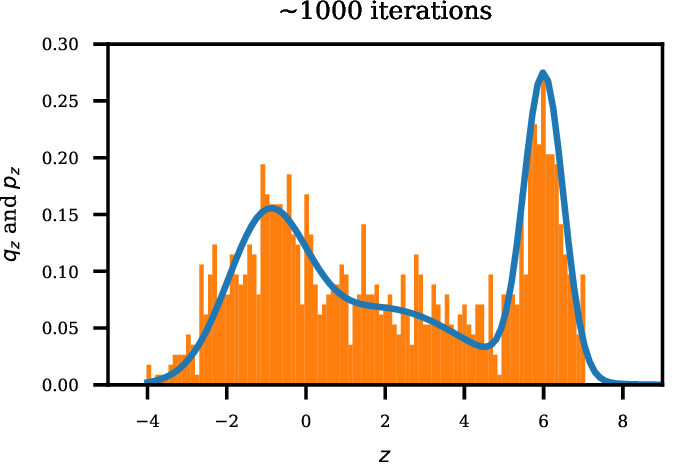}
  \caption{Mixture of three 1D Gaussians. The blue line indicates the target
    distribution, and the normalized histogram corresponds to generated values.\label{fig:toy1}}
  \end{subfigure}
  \vspace{1em}

  \begin{subfigure}{.8\textwidth}\centering
  \includegraphics[width=.49\textwidth]{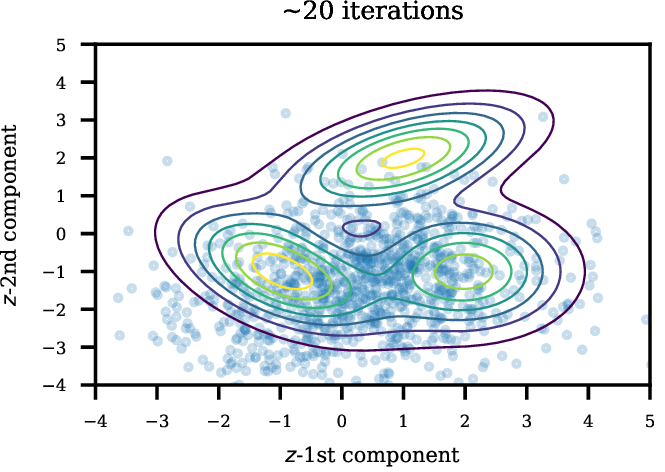}
  \hfill
  \includegraphics[width=.49\textwidth]{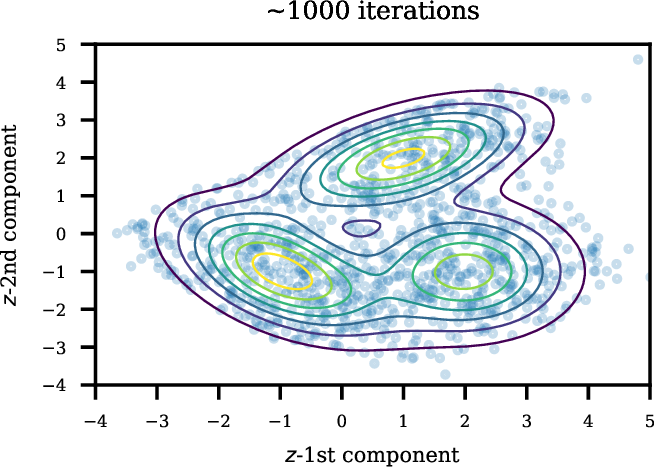}
  \caption{Mixture of three 2D Gaussians. The contour lines indicate the target
    distribution, and the scattered points correspond to generated values.\label{fig:toy2}}
  \end{subfigure}

  \caption{Results of $I_\phi$ trained to generate mixture of Gaussians. \label{fig:toy}}
\end{figure}

The proposed method described in~\Cref{sec:methodology} can be used to train a general \emph{neural sampler}.
In this side section, we perform a simple sanity check by assessing the method on a toy problem where we train neural networks to sample mixture of Gaussians.
Concretely, we train fully connected neural networks $I_\phi\colon\RR^{n_w}\to\RR^{n_z}$ to sample simple 1D and 2D mixture of Gaussians,
with $n_z=n_w=1$ in the 1D case, and $n_z=n_w=2$ in the 2D case. The source distribution $p_w$ is the standard normal in both cases.
Results are summarized in~\Cref{fig:toy}.

The first example (\Cref{fig:toy1}) is a mixture of three 1D Gaussians, with centers $\mu_1=-1$, $\mu_2=2$ and $\mu_3=6$, and standard deviations $\sigma_1=1, \sigma_2=2, \sigma_3=0.5$, respectively. The density of the Gaussian mixture is indicated along with a histogram for $1000$ points generated by the neural network at an early stage of the training (100 iterations), and at convergence (1000 iterations).
The second example (\Cref{fig:toy2}) is a mixture of three 2D Gaussians, with centers
$\mu_1=(-1,-1)$, $\mu_2=(1,2)$ and $\mu_3=(2,-1)$, and covariances
$\Sigma_1=\bigl(\begin{smallmatrix} 1&-0.5 \\ -0.5&1 \end{smallmatrix} \bigr)$,
$\Sigma_2=\bigl(\begin{smallmatrix} 1.5&0.6 \\ 0.6&0.8 \end{smallmatrix} \bigr)$, and
$\Sigma_3=\bigl(\begin{smallmatrix} 1&0 \\ 0&1 \end{smallmatrix} \bigr)$, respectively.
We plot the contour lines of the density of the Gaussian mixture. We also show
a scatter-plot of 4000 points generated by the neural network at an early stage of the training (20 iterations), and at convergence (1000 iterations).
In both test cases, we can verify that the neural network effectively learns to
transport points from the standard normal distribution to the mixture of
Gaussians.

\section{Comparison to related work based on inpainting} \label{sec:inpainting}
In image processing, image inpainting is used to fill incomplete images or replace a subregion of an image (e.g. a face with eyes covered). The recent GAN-based inpainting technique employed in~\citep{dupont2018generating,mosser2018conditioning} uses an optimization approach with the following loss:

\begin{equation}\label{eq:yeh}
  \manL(z) = \|G(z)_{\mathrm{obs}}-d_{\mathrm{obs}}\|^2 + \lambda \log(1-D(G(z)))
\end{equation}
The second term in this loss function is referred to as the \emph{perceptual loss} and
is the same second term in the GAN loss in~\Cref{eq:ganloss}, which is
the classification score on synthetic realizations.
Compare \Cref{eq:yeh} with~\Cref{eq:lossfn}: While our Bayesian posterior uses a simple Gaussian prior, the prior in~\Cref{eq:yeh} (the perceptual loss) involves the discriminator $D$ used during the GAN training.
We argue that the Gaussian prior can be equally effective, as long as the GAN training has converged successfully:
If $G$ and $D$ are at convergence, then $G(\zz)$ always produces plausible realizations for $\zz\sim p_z$ where $p_z$ is the chosen latent distribution, and $D$ is $1/2$ for all realizations of $G(\zz)$.
In such scenario, the perceptual loss should then act as a regularization term that drives $z$ towards regions of high density of the latent distribution $p_z$, therefore having a similar effect to using $p_z$ as the prior.

For example, let us consider $\zz\sim\manU [0,1]$ and $\yy\sim\manU [1,3]$.
An optimal generator would be $G(z)=2z+1$ and an optimal discriminator $D(y)=1/2$ for $y\in[1,3]$ and $D(y)=0$ otherwise.
Then $D(G(z))=1/2$ for $z\in[0,1]$, and $D(G(z))=0$ otherwise, which is precisely the density function of $\zz\sim\manU [0,1]$ scaled by $1/2$.
Therefore, in this example the perceptual loss and $p_z$ as prior would have the same effect.
Nevertheless, in practice the perceptual loss can be very useful when $G$ and $D$ are not exactly optimal and there exist bad realizations from $G$. In that case, the perceptual loss can help the optimization to find good solutions. %
In our work, we found our Gaussian prior to be sufficient while removing a layer of complexity in the optimization.

% Finally, we also note that~\citep{yeh2016semantic} explored both L1 and L2 norms for the likelihood term of~\Cref{eq:yeh}, with L1 resulting in a term $\propto \exp(-\frac{1}{\lambda}\|d(z) - d_{\mathrm{obs}} \|)$.

\section{Convolutional neural networks}\label{sec:cnn}

\begin{figure}\centering
	\begin{subfigure}{\textwidth}\centering
		\includegraphics[width=.6\textwidth]{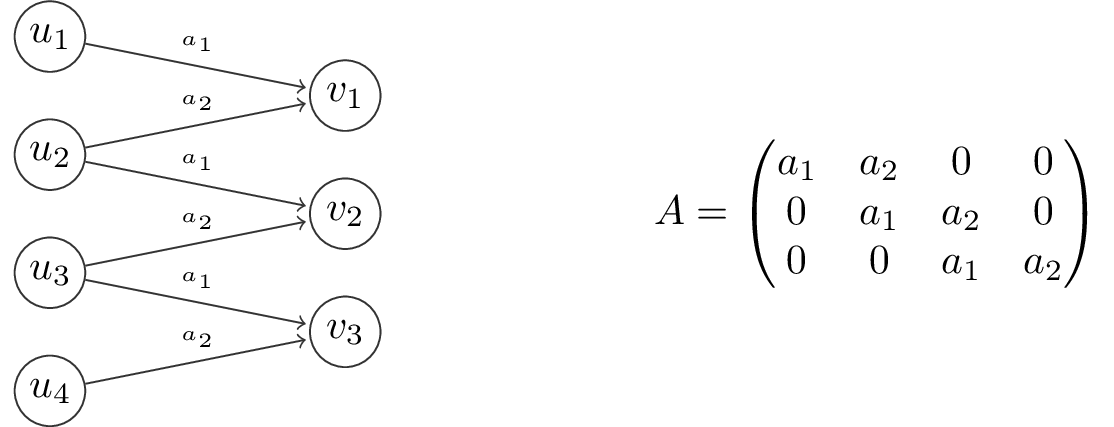}
		\caption{A convolutional layer.\label{fig:cnn}}
	\end{subfigure}\vspace{1em}
	\begin{subfigure}{\textwidth}\centering
		\includegraphics[width=.6\textwidth]{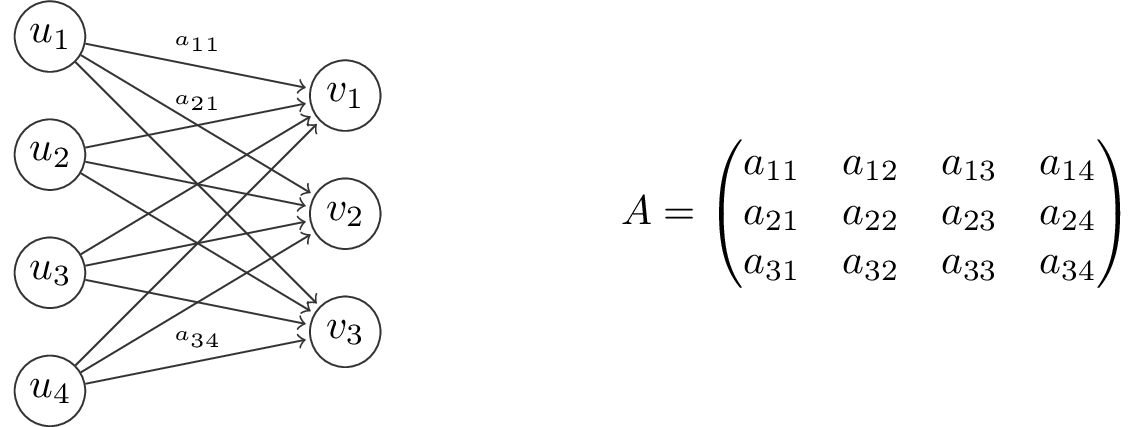}
		\caption{A fully connected layer.\label{fig:fc}}
	\end{subfigure}
	\caption{Neural network layers and respective transformation matrices.}
\end{figure}

We provide a very brief description by example of convolutional neural networks. See~\citep{fukushima1982neocognitron,lecun1989backpropagation} for further details or~\citep{dumoulin2016guide} for a more practical treatment.
Let $u = (u_1,u_2,u_3,u_4)$ and $a = (a_1,a_2)$. Let us call $a$ a \emph{filter}.
To convolve the filter $a$ on $u$ is to compute the output vector $v$ with components $v_i = u_ia_1 + u_{i+1}a_2$ for $i=1,\cdots,3$. The operation is illustrated as a neural network layer in~\Cref{fig:cnn}.
In this example, the convolution has a stride of 1 (at which the filter is swept), but in general it can be any positive integer.

We also show the matrix $A$ associated with this operation -- it is easy to verify that $v = Au$.
We see that the associated matrix is sparse and diagonal-constant, which is the appeal of using convolutional layers.
This structural constraint achieves two things: it drastically reduces the number of free weights, and it does so by assuming a locality prior. This locality prior turns out to be useful in practice, since nearby events in natural phenomena (natural images, speech, text, etc.) tend to be correlated.

Compare the convolutional layer with the fully connected layer shown in~\Cref{fig:fc}: In the fully connected case, the associated matrix is dense, resulting in 12 free weights whereas the convolution layer has only 2 for the same layer sizes. This difference is greatly amplified in practice where inputs/outputs are large (e.g. images), making convolutional layers a much more efficient architecture.
Note that in practice we use \emph{deep} architectures, i.e. several stacks of convolutional layers, therefore the full connectivity can be recovered if necessary, although now with an embedded locality prior along with a hierarchy in the influence of the weights.

Note that the example considered above would always result in a smaller output vector size. If the opposite effect is desired, a simple solution is to transpose the matrix $A$. For this reason, this operation is called a transposed convolution. Several stacks of transposed convolutions are typically used in generators and decoders to upsample the small latent vector to the full-size output image.
In classifier neural networks, normal convolutions are used instead to downsample the large image to a single number indicating a probability.

Our brief description can be readily extended to 2D and 3D arrays with corresponding multidimensional filters. For example, for a 2D input the filters are of rectangular shape and can be swept horizontally and vertically.
See~\citep{dumoulin2016guide} for further practical details.

\section{Computational complexity}\label{sec:computational}
Let $N$ denote the dataset size and $d$ the dimension of each realization.
Fast PCA methods based on singular value decomposition can achieve a
complexity of $\manO(N^2d)$. This complexity is favorable in geology where
$N\ll d$, i.e. we have a small number of very large realizations (although it
still grows quadratically with the number of realizations).
For our present method, reporting the computational complexity
is less straightforward since it is highly problem-dependent. To
illustrate the difficulties, we discuss in the following the computational
complexity of a classifier neural network -- similar arguments apply to
encoders, generators and decoders.

Computing the computational complexity of neural network models is cumbersome
since it fully depends on the architecture, which in turn depends on the
\emph{learning difficulty} of the problem at hand.
For example, in the simple case that the dataset is linearly separable, a
classifier neural network of the form $f(x)=\sigma{(w^Tx+b)}$, with $w,b$ to be
determined, is enough to correctly classify all points of the dataset.
The evaluation cost of this neural network is simply $\manO(d)$, hence the
training cost is $\manO(Td)$ (when using stochastic gradient descent as
normally done), where $T$ is the number of update iterations.
Note that this expression does not depend on $N$, although in practice $T$ is
at most linear in $N$, e.g. when performing multiple passes through the
dataset until convergence, but note that the training can also converge even
before a single pass through the dataset (which happens on massive
datasets).
Hence, neural networks are very favorable in the big data setting, i.e. when $N$
is very large.

The estimated evaluation cost of $\manO(d)$ is overly optimistic since in practice we use \emph{deep}
architectures to deal with complex datasets that are not linearly separable.
If the architecture is instead $f(x)=\sigma(A_l(\cdots\sigma(A_2(\sigma(A_1x+b_1)+b_2))\cdots)+b_l)$,
where each $A_i$ is a $d\times d$ matrix,
then the evaluation cost of this architecture is roughly $\manO(d^2)$ (we
omit the number of layers $l$ since this is a constant factor and $l\ll d$).
However, this estimate is now overly pessimistic: First, in practice the $A_i$
are not shape-preserving, instead they decrease very quickly in size while
exponentially compressing the input (e.g. $A_1$ is of size $d\times
\frac{d}{2}$, $A_2$ is of size $\frac{d}{2}\times
\frac{d}{4}$, etc.).
Second, the matrices $A_i$ are rarely full since convolutional layers are
used instead (see~\Cref{sec:cnn}), resulting in very sparse matrices that are
several orders of magnitude lighter.
Modern architectures use several stacks of exponentially decreasing
convolutional layers, while fully connected layers are avoided or used only sparingly
(and for small inputs/outputs).
The overall effect is a drastic reduction in the computational complexity,
from $\manO(d^2)$ to $\manO(kd)$ where $k$ is a factor that is determined by
the architecture. The corresponding training complexity is then $\manO(Tkd)$.
Note that although $k < d$ in practice, $k$ can still be sizable. On the
other hand, $k=1$ is also possible as just mentioned.
Ultimately, $k$ will depend on the learning difficulty of the problem.
In most models encountered in the literature, $k$ grows sublinearly with $d$.

Perhaps more importantly is the human time, rather than computational time,
that is involved in optimizing the dozens of hyperparameters -- in particular
the architecture design -- for which automation is currently limited. As
mentioned before, designing the architecture is heavily based on experience,
heuristics, and experimentation which incur high costs in terms of
engineering time. The justification of such costs will ultimately depend on
the lifespan of the model, since the model needs to be constructed only once but can be deployed
for a long time (e.g. history matching) or virtually indefinitely (e.g. most
applications in internet companies such as recommender systems, visual and
voice recognition, language translation, etc.). Automatic architecture search
is an ongoing area of research (see e.g.~\citep{shahriari2016taking,zoph2016neural} and
references therein).

\end{document}